\documentclass[11pt]{article}

\usepackage[final]{acl}

\usepackage{times}
\usepackage{latexsym}
\usepackage[T1]{fontenc}
\usepackage[utf8]{inputenc}
\usepackage{microtype}
\usepackage{graphicx}
\usepackage{algorithm}
\usepackage{algpseudocode}
\usepackage{algorithmicx}
\usepackage{amsmath}
\usepackage{amssymb}
\usepackage{amsthm}
\usepackage{rotating}
\usepackage{tabularx}
\usepackage{booktabs}
\usepackage{tikz}
\usetikzlibrary{positioning}
\usepackage{xcolor}
\usepackage{adjustbox}
\usepackage{multirow}
\usepackage{enumitem}
\usepackage{makecell}

\usepackage{bm}
\usepackage[breakable,skins]{tcolorbox}
\usepackage{fancyvrb}
\newcounter{boxA}

\title{EmoDistill: Offline Emotion Skill Distillation for \\Language Model Agents in Adversarial Negotiation}

\author{
  \textbf{
  Yunbo Long$^{1}$\thanks{Equal contribution}
  \quad
  Haolang Zhao$^{1}$\footnotemark[1]
  \quad
  Lukas Beckenbauer$^{2}$
  } \\
  \textbf{
  Liming Xu$^{1,3}$
  \quad
  Alexandra Brintrup$^{1,4}$\thanks{Corresponding author}
  } \\
  $^{1}$University of Cambridge \quad
  $^{2}$Technical University of Munich \\
  $^{3}$Exiger LLC \quad
  $^{4}$The Alan Turing Institute \\
  \texttt{\{yl892,hz496,lx249,ab702\}@cam.ac.uk}
  \quad
  \texttt{lukas.beckenbauer@tum.de}
}

\begin{document}
\maketitle

\begin{abstract}
Post-trained LLMs are often optimized to produce helpful, polite, and accommodating responses.
In adversarial negotiation, however, such behavior can become a vulnerability: emotionally framed language may influence an agent's bargaining decisions in ways that conflict with its user's objectives.
We therefore introduce \textbf{EmoDistill}, an offline framework for distilling emotional negotiation skills from LLM--LLM interactions into smaller language-model agents.
Here, an emotional negotiation skill is a state-conditioned behavior that determines which explicit emotion to invoke in a bargaining state and how to realize that emotion as an effective negotiation utterance.
EmoDistill learns these two components separately: an Implicit Q-Learning (IQL) selector learns which emotion to express in each bargaining state, while a LoRA-adapted 7B policy learns the emotion-conditioned expression through Supervised Fine-Tuning (SFT) and Judge Policy Optimization (JPO).
Across four emotion-sensitive negotiation domains, the full EmoDistill policy achieves competitive Utility and improves over vanilla and IQL-only baselines in most settings.
Emotion-free ablations show that removing the explicit emotion channel substantially reduces overall negotiation utility, while transfer experiments reveal partial, domain-dependent transfer and robustness to unseen LLM counterparties.
The code is available at \url{https://github.com/Yunbo-max/EmoDistill}.
\end{abstract} 

\section{Introduction}

\begin{figure}[t!]
    \centering
    \includegraphics[width=0.95\columnwidth]{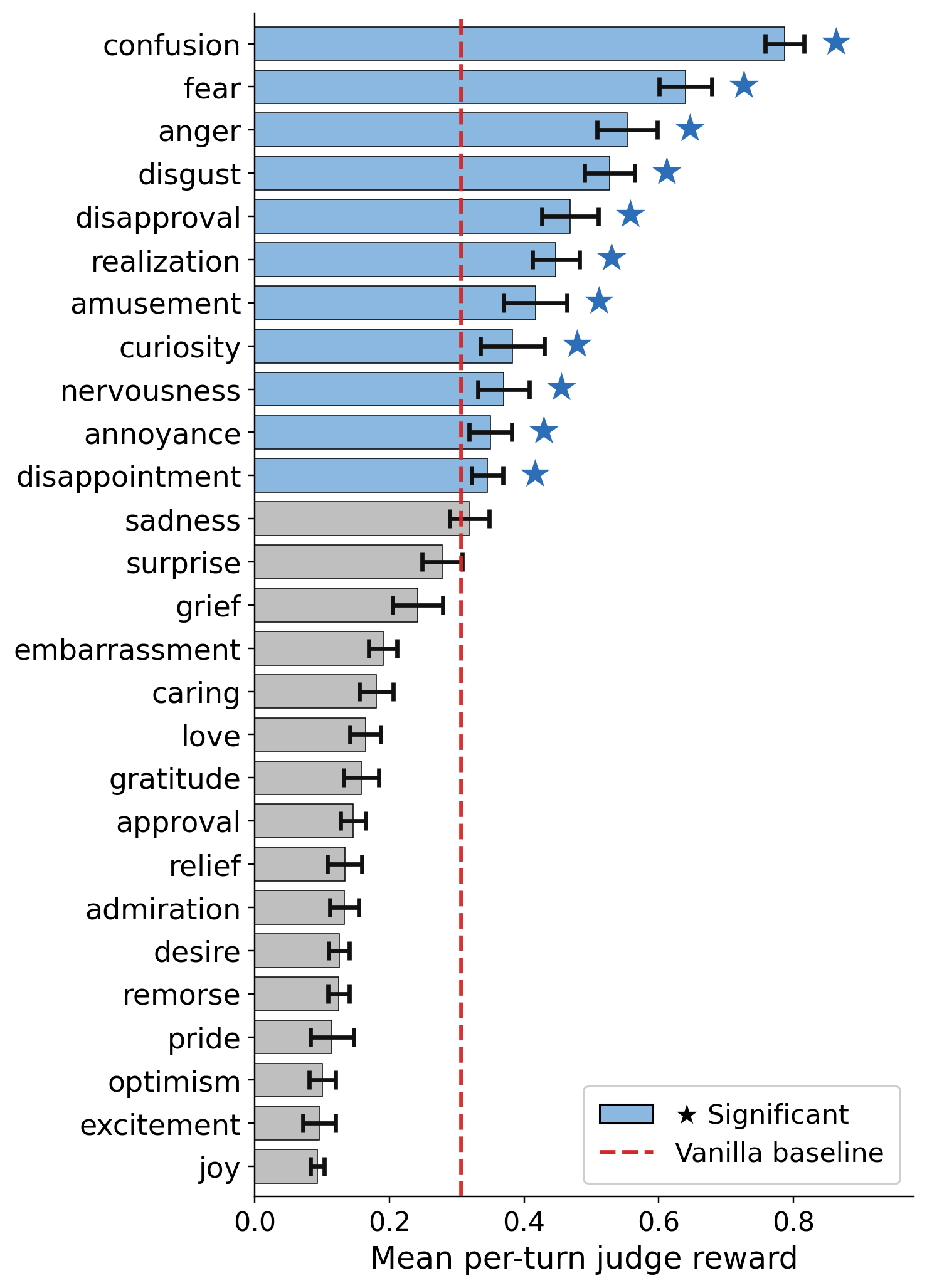}
    \caption{Single-emotion prompting effects on CRAD Debt Negotiation. GoEmotions labels are ranked by mean per-turn judge reward ($\pm95\%$ CI), with the vanilla baseline shown as a dashed line. This analysis tests whether explicit emotion prompts systematically affect negotiation behavior.}
    \label{fig:emotion_subset}
\end{figure}

\begin{figure*}[t]
    \centering
    \includegraphics[width=\textwidth]{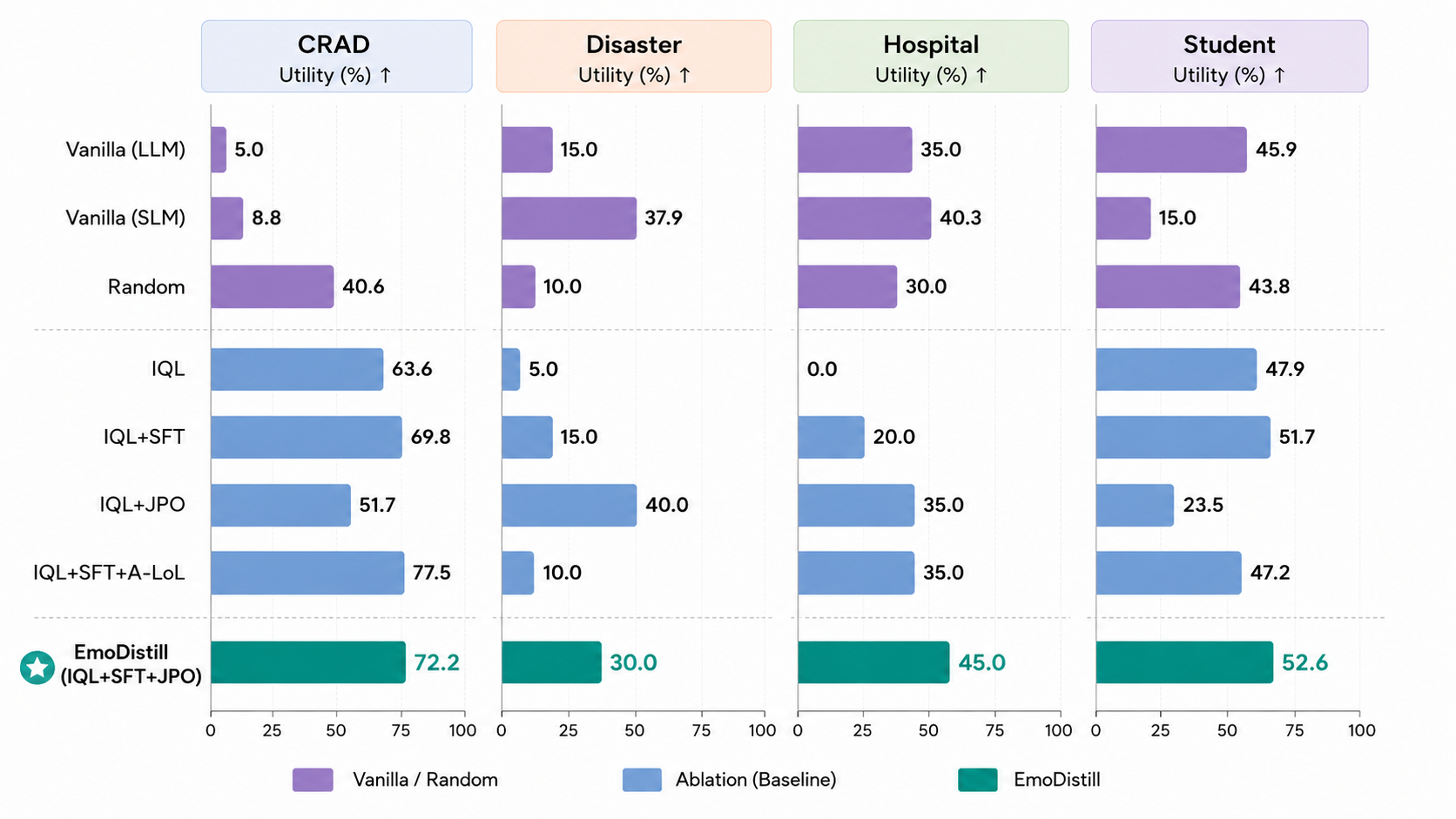}
    \caption{
Utility comparison across four negotiation domains.
A-LoL uses the same IQL selector and LoRA-SFT initialization as EmoDistill.
EmoDistill achieves the competitive performance across four domains.
}
\label{fig:main_results}
\end{figure*}

Modern large language models (LLMs) are extensively post-trained through RLHF~\citep{kasbouya2025emotional}, DPO~\citep{gao2025emo}, and instruction tuning to produce helpful, polite, and accommodating responses.
While these properties are desirable for general-purpose assistants, they can become liabilities when the same models are deployed as autonomous agents in strategic or adversarial interactions.
This issue is particularly relevant for tool-calling agents that can take consequential actions on behalf of users, such as transferring money, booking travel, purchasing products, scheduling meetings, or handling customer support~\citep{lin2024large,abbasiantaeb2024let,hu2025trustless}.
Many of these interactions involve negotiation over price, time, priority, refunds, or deadlines.
When both parties are language-model agents and no human directly intervenes in each decision, emotional framing becomes another channel through which one agent may influence another.
Emotion can therefore act both as a source of strategic vulnerability and, when explicitly modeled, as a controllable bargaining variable.

To test whether emotion meaningfully affects negotiation behavior rather than merely changing surface wording, we first conduct a controlled single-emotion prompting study on CRAD using the 28 GoEmotions labels~\citep{demszky2020goemotions}.
For each emotion, we evaluate the LLM negotiator on the same $20$ held-out scenarios across $20$ sampled runs and measure the mean per-turn judge reward.
As shown in Figure~\ref{fig:emotion_subset}, several emotion prompts produce systematic and statistically significant changes relative to the vanilla prompt.
Instead, a learned policy must determine which emotion is appropriate for the current bargaining state and how to express it.
This motivates treating emotion as an explicit action channel rather than a fixed stylistic prompt.

A negotiator therefore faces two distinct decisions.
First, it must decide \emph{which emotion} is strategically appropriate in the current dialogue state.
Second, it must decide \emph{how to realize} that emotion as an effective bargaining utterance.
These decisions are not equivalent.
Selecting anger does not by itself determine whether the resulting response will contain a credible rejection, an appropriate numerical anchor, or an unjustified ultimatum.
Similarly, fear may communicate useful urgency in one state but signal unnecessary weakness in another.
The strategic value of emotion therefore depends on both state-dependent selection and state-grounded language realization.

Existing emotion-aware negotiation methods address mainly the first problem by modeling emotion as a dynamic decision variable~\citep{long2026eq,long2025emodebt,long2025evoemo}.
However, they largely optimize \emph{which} emotion to express while leaving the underlying utterance generator fixed.
A strategically appropriate emotion may consequently still be realized as vague politeness, repetitive language, premature concession, or weak justification.
The missing component is \emph{strategic emotional expression}: learning how a selected emotion should be converted into a concrete bargaining move under the current negotiation state.

Learning both components directly with online reinforcement learning is costly in LLM agent-to-agent negotiation.
Each policy update would require new stochastic multi-turn interactions with an API-served counterparty, followed by repeated reward annotation.
Such trajectories are expensive to regenerate and cannot be efficiently reused across different optimization stages.
Furthermore, terminal outcomes provide only coarse feedback: they reveal whether a complete negotiation succeeded, but not which particular utterances strengthened or weakened the trajectory.
Pure on-policy methods such as PPO~\citep{schulman2017proximal} therefore require substantially more live interaction than a reusable offline approach.

We address this problem through \emph{offline emotion skill distillation}.
An emotional negotiation skill is not an emotion label, a fixed ``act angry'' instruction, or a hand-written prompt template.
Instead, it is a learned, state-conditioned bargaining behavior that connects
(i) the dialogue state $s_t$,
(ii) an explicit emotion action $e_t$,
and
(iii) the concrete negotiation utterance $u_t$ that realizes that action.
The offline LLM--LLM sweep provides many state--emotion--utterance instances together with trajectory-level and turn-level feedback.
Individual tuples are training instances of the skill; the distilled skill itself is the learned mapping from negotiation state to emotion choice and from emotion choice to bargaining language.

EmoDistill distills this behavior at two complementary levels.
At the strategic level, an IQL selector learns
\[
\pi_\phi(e_t\mid s_t),
\]
which determines \emph{which emotion} to invoke in a given bargaining state using objective trajectory returns.
At the language level, a LoRA-adapted SLM learns
\[
\pi_\theta(u_t\mid s_t,e_t),
\]
which determines \emph{how to express} the selected emotion as a concrete negotiation utterance.
LoRA-SFT initializes this expression policy from high-quality offline demonstrations, while Judge Policy Optimization (JPO) further refines it using normalized turn-level judge feedback.
Thus, EmoDistill does not distill emotion labels themselves; it distills the state-dependent mapping from negotiation context to emotional strategy and from emotional strategy to language.

The process is fully offline.
After the LLM--LLM negotiation sweep has been collected and annotated once, the same fixed trajectories are reused throughout training.
IQL learns emotion selection from objective trajectory returns, SFT selects and imitates high-quality utterances, and JPO refines expression using turn-level advantages.
No new student--counterparty negotiation rollouts are required during optimization.
This makes the expensive agent-to-agent interactions reusable while providing supervision at multiple temporal granularities.

We therefore propose \textbf{EmoDistill}, an offline framework for transferring state-conditioned emotional negotiation behavior from LLM--LLM interactions into a smaller negotiator.
By explicitly separating emotion selection from emotion expression, EmoDistill allows a 7B SLM to learn both when an emotional strategy is useful and how that strategy should be realized in language.

Our main contributions are:
\begin{itemize}[leftmargin=*]

    \item We introduce \textbf{EmoDistill}, an offline framework for distilling state-conditioned emotional negotiation skills from fixed LLM--LLM interactions into a smaller language-model agent.
    The framework learns both a state-dependent emotion-selection policy and an emotion-conditioned language policy without requiring online negotiation rollouts during optimization.

    \item We empirically separate the roles of emotion selection and emotion expression.
    IQL learns which of the 28 explicit emotion actions to invoke in a bargaining state, while LoRA-SFT learns how to realize the selected emotion as a concrete negotiation utterance.
    Emotion-free ablations show that removing this explicit emotion-control channel substantially reduces overall negotiation utility.

    \item We propose \textbf{Judge Policy Optimization} (JPO), an offline refinement stage that uses normalized turn-level LLM-judge scores as advantages to improve the SFT-trained expression policy under clipping and KL regularization.
    This provides utterance-level supervision that complements the trajectory-level signal used for emotion selection.

\end{itemize}

\begin{figure*}[t]
    \centering
    \includegraphics[width=\textwidth]{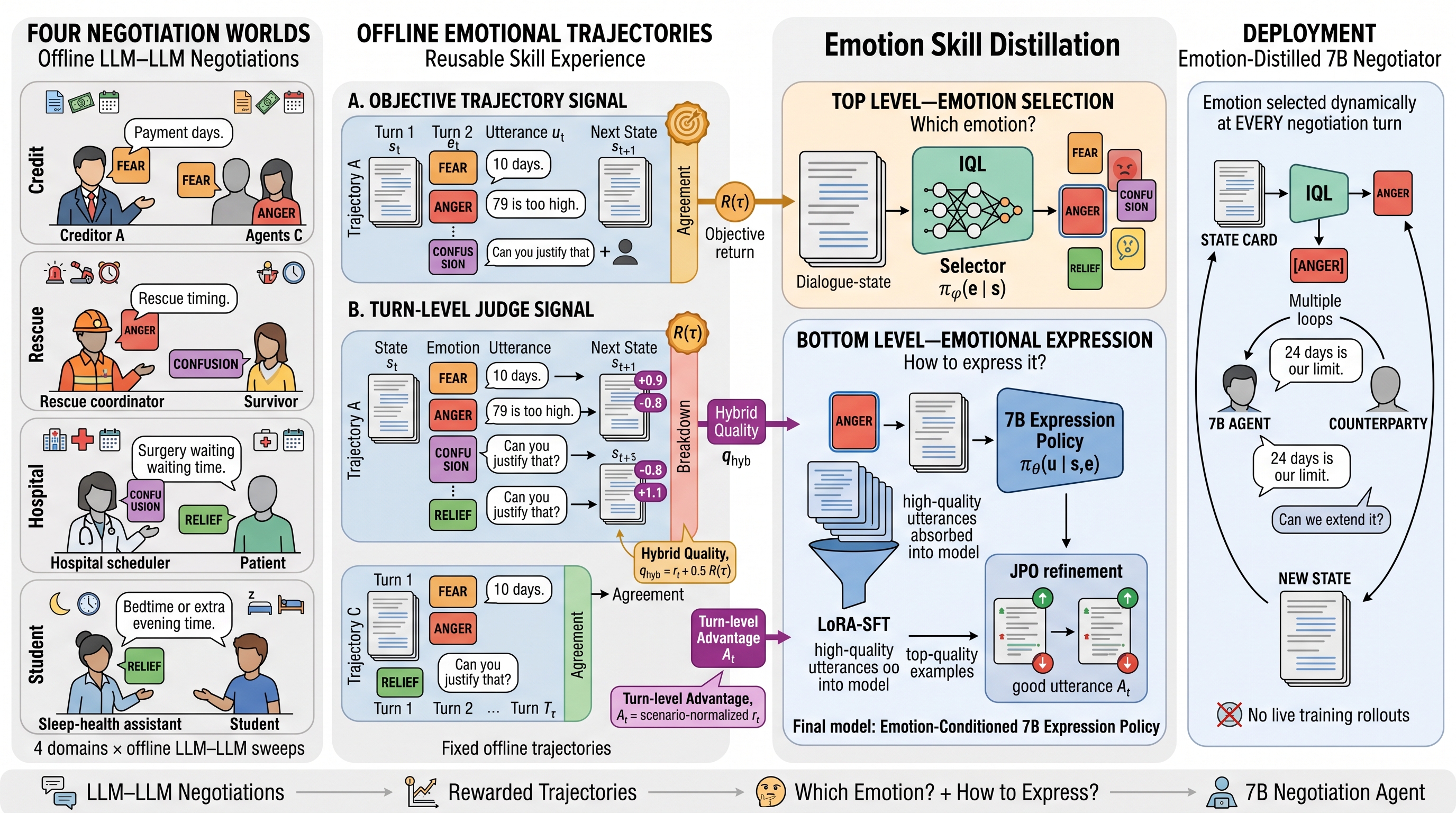}
    \caption{Overview of EmoDistill.
    A fixed offline LLM--LLM negotiation sweep provides state--emotion--utterance trajectories together with objective trajectory returns and turn-level judge scores. EmoDistill distills two complementary skills: an IQL selector learns \emph{which emotion} to invoke in each bargaining state, while a LoRA-adapted 7B expression policy learns \emph{how to express} the selected emotion through a concrete negotiation utterance.}
\label{fig:workflow}
    
    \label{fig:workflow}
\end{figure*}

\section{Related Work}
\label{sec:related_work}

\paragraph{Emotion in agent-to-agent negotiation.}
Modern LLM agents are trained on human-authored text and dialogue, and therefore inherit affective and pragmatic patterns such as politeness, empathy, and concession framing. When these models negotiate with other agents, such patterns become part of the bargaining interface. Since prior work shows that emotions can serve as dynamic strategic instruments in negotiation~\citep{huang2024personality,griessmair2015emotions,olekalns2014feeling}, we treat emotion as a controllable action channel in LLM-based agent-to-agent bargaining. Recent LLM-based negotiation systems have begun treating emotion as a meaningful variable, but typically as an input rather than an optimized output. AgreeMate~\citep{chatterjee2024agreemate} and ACE~\citep{shea2024ace} use emotion-aware reasoning, while EQ-Negotiator~\citep{long2026eq} combines emotion sensing with Hidden-Markov reasoning. EmoDebt~\citep{long2025emodebt} and EvoEmo~\citep{long2025evoemo} go further by treating emotion as a sequential decision variable, optimized via Bayesian optimization and evolutionary search respectively. However, all of these optimize \emph{which} emotion to express while leaving the utterance generator fixed; a selected emotion may still be realized through vague politeness or premature concession. EmoDistill addresses this gap by jointly distilling emotion selection and emotional expression into a smaller model.
 
\paragraph{Decoupling strategy from expression.}
\citet{he2018decoupling} first proposed decoupling high-level coarse dialogue acts from utterance generation in negotiation, observing that end-to-end RL tends to collapse to degenerate solutions such as repetitive utterances or meaningless concessions that exploit the reward signal. This idea has been extended to cooperative emotional-support dialogue: EmoDynamiX~\citep{wan2025emodynamix} decouples strategy prediction from generation via heterogeneous graph modeling, and DecoupledESC~\citep{zhang2025decoupledesc} uses strategy-response decoupled DPO to mitigate preference bias. Our setting is adversarial rather than cooperative, and the strategic axis is \emph{emotion} rather than price-level acts; these differences require different training signals and a different decoupling mechanism. EmoDistill decouples emotion selection (an offline IQL selector) from emotional expression (a LoRA-adapted SLM generator), distilling both into a 7B student rather than coordinating a frozen LLM with a retrieval module.

\paragraph{Offline distillation with LLM-judge signals.}
\label{subsec:related_distill}
LLM judges provide scalable supervision when human labels or online rollouts are expensive. 
RLAIF~\citep{lee2023rlaif,bai2022constitutional} uses AI-generated preferences for sequence-level alignment, while process reward models~\citep{lightman2023verify} provide step-level feedback mainly for reasoning tasks. 
A-LoL~\citep{baheti2024leftover} is related as an offline advantage-based method for language-model refinement: it treats the entire generated response as one action and trains on positive-advantage examples. 
However, multi-turn negotiation requires a different credit-assignment structure. 
In EmoDistill, each focal-agent turn is a reward-annotated emotional bargaining move, and JPO refines the expression policy using scenario-normalized turn-level judge advantages. 
This lets the model learn which emotional utterances move the bargaining trajectory toward or away from the focal target, rather than only amplifying sequence-level positive examples. 
We compare A-LoL and JPO refinement in Appendix~\ref{app:alol_baseline}.


\section{EmoDistill}
\label{sec:method}

Figure~\ref{fig:workflow} provides an overview of the pipeline showing an offline pipeline with three stages. 
We first construct an LLM-vs-LLM negotiation dataset and attach two complementary signals to each offline trajectory: a dense per-turn LLM-judge score for each focal-agent utterance and an outcome-shaped trajectory return computed from observed bargaining dynamics and terminal agreement. 
The same offline sweep is reused across all training stages: IQL uses the outcome-shaped return for emotion selection (\S\ref{sec:iql}), LoRA-SFT uses a hybrid judge--outcome filter for demonstration selection, and JPO uses dense judge-derived advantages for utterance-level policy improvement (\S\ref{sec:expression}). 
Section~\ref{sec:reward} formalizes this stage-wise signal design. 
The full EmoDistill policy is reported in experiments as \textbf{IQL+SFT+JPO}: IQL selects which emotion to invoke, LoRA-SFT initializes how that skill is expressed, and JPO refines the utterance generator with dense judge-derived advantages.

\subsection{Offline Trajectory Dataset and Judge Annotation}
\label{sec:dataset}

For each domain (CRAD, Disaster Rescue, Hospital Surgery, Student Sleep), we collect $N{=}80$ training scenarios $\times$ $M{=}100$ random emotion-sequence rollouts, yielding an offline dataset $\mathcal{D}$ of 8000 trajectories per domain. 
Each rollout samples emotions from the full action vocabulary $\mathcal{E}$ (28 GoEmotions labels). 
At each focal-agent turn, $\mathcal{D}$ records $z_t=(s_t,e_t,u_t,r_t,s_{t+1})$, where $s_t$ is the dialogue state, $e_t \in \mathcal{E}$ is the emotion action, $u_t$ is the focal utterance, $r_t$ is the judge-assigned per-turn reward, and $s_{t+1}$ is the next state after the counterparty responds. 
Each high-reward instance provides a training example of an \emph{emotional negotiation skill}: a state-grounded emotional action realized through a concrete bargaining utterance.
The distilled skill itself is the learned state-dependent mapping from negotiation state to emotion choice and from the selected emotion to bargaining language.
The full per-turn judge rubric prompt is given in Appendix~\ref{app:prompts_judge}. 
Dataset details, sweep construction, and the prompt interface are given in Appendices~\ref{app:dataset_details}, \ref{app:sweep_construction}, and~\ref{app:prompts}.

\subsection{Reward Design and Stage-wise Signal Use}
\label{sec:reward}

EmoDistill draws training signals from two complementary sources: a \textbf{per-turn LLM judge} that provides dense subjective evaluation of each focal utterance, and an \textbf{objective trajectory return} computed from observed bargaining dynamics and terminal agreement.

\paragraph{Subjective signal.}
A Qwen3.5-Plus judge scores each focal utterance against a metric-aligned rubric that rewards anchoring toward the focal target, concrete proposals, and scenario-grounded leverage, while penalizing capitulation, vagueness, repetition, poorly calibrated concessions, and breakdown-inducing behavior. 
We denote the raw per-turn score by $r_t$ and normalize within scenario:
\begin{equation}
A_t = \frac{r_t - \mu_\text{scen}}{\sigma_\text{scen} + \epsilon}.
\label{eq:advantage}
\end{equation}

\paragraph{Objective signal.}
The outcome-shaped reward $R(\tau)$ rewards the focal agent for shifting the bargaining gap in its favor:
\begin{equation}
R(\tau) = \underbrace{\sum_{t=1}^{T_\tau} w(t)\bigl(\Delta^{\text{ctp}}_t - \Delta^{\text{foc}}_t\bigr)}_\text{step shaping}
+ \underbrace{R^\text{term}(\tau)}_\text{agreement bonus}.
\end{equation}
The step shaping credits turns where the counterparty concedes more than the focal agent: $\Delta^{\text{ctp}}_t$ is the counterparty's per-turn move toward the focal target (positive when they close the gap), and $\Delta^{\text{foc}}_t$ is the focal agent's own retreat (positive when they move away from their target), both normalized by the initial anchor-to-target gap. 
The terminal anchor $R^\text{term}(\tau){=}{+}2$ for reached agreement and $-2$ for breakdown. 
The linear time-decay $w(t){=}\max(0, \min(1, 1{-}t/T_\text{max}))$ down-weights late-turn concessions as an implicit length penalty. 
$R(\tau)$ uses no LLM-judge signal; the full formulation and reward-variant definitions are in Appendix~\ref{app:reward_design}.

\paragraph{Stage-wise signal use.}
The two signals are repurposed across stages. 
\textbf{(i) IQL} (\S\ref{sec:iql}) uses the objective $R(\tau)$ as a Bellman-propagated terminal reward, so the selector is rewarded for emotion sequences that actually close the bargaining gap rather than those that merely sound persuasive. 
\textbf{(ii) LoRA-SFT} (\S\ref{sec:expression}) uses a hybrid filter combining $r_t$ and $R(\tau)$ to select demonstrations that are both locally well-formed and globally productive. 
\textbf{(iii) JPO} (\S\ref{sec:expression}) uses the subjective per-turn $A_t$ for clipped offline policy improvement, enabling credit assignment at the level of individual emotional expressions. 
In Sec.~\ref{sec:experimental_setting} we ablate three reward variants per stage, differing in how the signal is distributed across turns: \emph{outcome-shaped} (objective, \emph{sparse}: $R(\tau)$ only at trajectory end, propagated via Bellman backups), \emph{episode-judge} (subjective, \emph{broadcast}: one dialogue-level judge score copied to every turn), and \emph{turn-judge} (subjective, \emph{dense}: an independent judge score per focal turn). 
SFT benefits from hybrid quality filtering (\emph{which demonstrations to imitate}), whereas JPO benefits from dense turn-level advantages (\emph{which expressions to upweight or downweight}).

\subsection{Emotion Selection with Offline IQL}
\label{sec:iql}

The selector treats $e_t \in \mathcal{E}$ as the action and $s_t$ as the state. 
We train Implicit Q-Learning (IQL) on $\mathcal{D}$, learning $Q(s, e)$ and $V(s)$ with the standard expectile objective, and extract an advantage-weighted selector with temperature $\beta_\text{AWR}$:
\begin{equation}
\pi_\phi(e \mid s) \propto \exp\!\big(\beta_\text{AWR} \cdot (Q(s,e) - V(s))\big).
\end{equation}
At inference, the selector samples an emotion for the current dialogue state, which is then inserted into the expression policy's prompt. 
IQL learns \emph{which} emotional skill to invoke; it does not update the utterance generator. 
Detailed selector objectives and pseudocode are provided in Appendices~\ref{app:compared_policies} and~\ref{app:algorithms}.

\subsection{Distilling Emotional Expression}
\label{sec:expression}

The IQL selector chooses the emotional skill, but the base SLM still needs to learn how to execute it in language. 
We train a LoRA adapter on Qwen2.5-7B-Instruct in two stages.

\paragraph{Stage 1: LoRA-SFT initialization.}
We score each turn in $\mathcal{D}$ by a hybrid quality function $q^\text{hyb}_t = r_t + \tfrac{1}{2}R(\tau)$, where the per-turn judge $r_t$ rewards locally well-formed expression and the trajectory return $R(\tau)$ rewards turns drawn from globally productive negotiations. 
We retain the top 25\% of $(s_t, e_t, u_t)$ tuples ranked by $q^\text{hyb}_t$ as demonstrations and train the LoRA adapter to generate $u_t$ conditioned on $(s_t, e_t)$ via token-level cross-entropy:
\begin{equation}
\mathcal{L}_\text{SFT} = - \sum_{k=1}^{|u_t|} \log \pi_\theta(u_{t,k} \mid s_t, e_t, u_{t,<k}).
\end{equation}

\paragraph{Stage 2: Judge Policy Optimization (JPO).}
Freezing the SFT adapter as $\pi_\text{ref}$, JPO applies an offline PPO-clipped surrogate with importance ratio $\rho_t = \pi_\theta(u_t \mid s_t, e_t)/\pi_\text{ref}(u_t \mid s_t, e_t)$ and a K3 KL anchor~\citep{schulman2020approximating} to $\pi_\text{ref}$. 
To control how strongly JPO moves away from low-value utterances, we replace the symmetric advantage $A_t$ from Eq.~\eqref{eq:advantage} with an asymmetric variant:
\begin{equation}
\widetilde A_t =
\begin{cases}
A_t, & A_t>0,\\
\kappa A_t, & A_t\le 0,
\end{cases}
\end{equation}
where $\kappa{\in}[0,1]$ controls the weight on negative-advantage samples. 
Smaller $\kappa$ preserves more of the SFT deal-closing prior by weakening the push away from low-value utterances; larger $\kappa$ applies stronger pressure against such utterances. 
The JPO objective is then
\begin{equation}
\begin{aligned}
\mathcal{L}_\text{JPO} = -\, &\mathbb{E}\big[\min\!\big(\rho_t \widetilde A_t,\; \text{clip}(\rho_t, 1{-}\varepsilon, 1{+}\varepsilon) \widetilde A_t\big)\big] \\
 &+ \lambda_\text{KL} \cdot \text{KL}_\text{K3}[\pi_\theta \,\|\, \pi_\text{ref}].
\end{aligned}
\end{equation}
Training hyperparameters, including LoRA configuration, $\varepsilon$, $\lambda_\text{KL}$, and the validation protocol for $\kappa$, are reported in Appendix~\ref{app:model_training_setup}.

\subsection{Variants and Emotion-Free Ablation}
\label{subsec:method_variants}

In the main tables, IQL+SFT+JPO denotes the full EmoDistill policy, combining IQL emotion selection, LoRA-SFT expression initialization, and JPO expression refinement. We evaluate three component variants. IQL uses only the learned emotion selector and pairs it with the frozen LLM generator, testing whether emotion selection alone can improve an LLM negotiator. IQL+SFT replaces the frozen LLM generator with a LoRA-adapted SLM expression policy trained by supervised fine-tuning. IQL+JPO removes the SFT warm start and tests direct judge-guided refinement of the SLM expression policy. 
We also evaluate an emotion-free diagnostic variant. 
The default EmoDistill policy is emotion-conditional: the selector samples $\hat e_t\sim\pi^{\mathrm{IQL}}_\phi(\cdot\mid s_t)$ and the expression policy generates $\hat u_t\sim\pi_\theta(\cdot\mid s_t,\hat e_t)$. 
In the emotion-free variant, the emotion block is removed during both training and inference, so the adapter directly generates $\hat u_t\sim\pi^{\mathrm{free}}_\theta(\cdot\mid s_t)$. 
This tests whether the LoRA adapter can internalize emotional strategy without an explicit emotion channel; the covariate-shift analysis is given in Appendix~\ref{app:emotionfree_covshift}.

\section{Experimental Setup}
\label{sec:experimental_setting}

\paragraph{Datasets.}
We evaluate on four negotiation domains: Credit Recovery (CRAD)~\citep{long2026eq}, Disaster Rescue, Hospital Surgery Scheduling, and Student Sleep Scheduling from EmoMAS~\citep{long2026emomas}. 
Each dataset contains $100$ scenarios, split into $80$ training and $20$ held-out test scenarios. 
For each training scenario, we sample $100$ random emotion-sequence rollouts from the full $|\mathcal{E}|=28$ vocabulary, yielding $8000$ offline trajectories per domain. 
The domains cover different roles, objectives, and preference directions; dataset details and sweep construction are in Appendices~\ref{app:dataset_details} and~\ref{app:sweep_construction}.

\paragraph{Compared methods.}
Qwen3.5-Plus~\citep{yang2025qwen3} is used for the default offline LLM-vs-LLM sweep and judge annotations, while DeepSeek-V3 \citep{deepseekai2025deepseekv3technicalreport} and GPT-4o
\citep{openai2024gpt4ocard} are additionally used for reward-labeler robustness and cross-counterparty evaluation.
In the IQL-only baseline, the selected emotion is inserted into the prompt of a frozen Qwen3.5-Plus model to be an IQL-guided LLM negotiator. In the distilled student methods, the same IQL selector conditions a Qwen2.5-7B-Instruct \citep{qwen2025qwen25technicalreport} focal SLM: IQL+SFT, IQL+JPO, and IQL+SFT+JPO differ only in how the SLM expression policy is adapted with LoRA. Vanilla LLM/SLM omit the selector, using Qwen3.5-Plus and Qwen2.5-7B-Instruct respectively; Random uses uniformly sampled emotion prompts. 

\paragraph{Training.}
All learned policies are trained from the same fixed offline sweep generated with Qwen3.5-Plus in an LLM-vs-LLM negotiation setup. 
By default, IQL uses the outcome-shaped trajectory return for emotion selection, LoRA-SFT uses a hybrid judge--outcome filter for demonstration selection, and JPO uses scenario-normalized per-turn judge advantages for utterance-level refinement. 
We set $\kappa=1$ for JPO unless otherwise stated. 
The training-signal ablation compares alternative reward variants for these stages. 
Reward definitions are given in Appendix~\ref{app:reward_design}, and model/training details are in Appendix~\ref{app:model_training_setup}.

\paragraph{Evaluation.}
Each method is evaluated on the same $20$ held-out scenarios per domain. 
We report success rate, \textbf{Outcomes}, \textbf{Utility}, and negotiation rounds. 
\textbf{Outcomes} averages normalized savings over successful negotiations, whereas \textbf{Utility} averages over all scenarios and assigns $0$ to failures. 
The role-neutral savings formula and aggregation rules are provided in Appendix~\ref{app:evaluation_protocol}.

\begin{table*}[t]
\centering
\setlength{\tabcolsep}{3pt}
\setlength{\cmidrulewidth}{0.2pt}
\setlength{\aboverulesep}{1.2pt}
\setlength{\belowrulesep}{1.2pt}
\renewcommand{\arraystretch}{1.05}

\begingroup
\fontsize{9.2}{10.2}\selectfont
\resizebox{\textwidth}{!}{%
\begin{tabular}{l|ccc|ccc|ccc|ccc}
\toprule
& \multicolumn{3}{c|}{\textbf{CRAD}}
& \multicolumn{3}{c|}{\textbf{Disaster}}
& \multicolumn{3}{c|}{\textbf{Hospital}}
& \multicolumn{3}{c}{\textbf{Student}} \\
\textbf{Method}
& \textbf{Suc.} & \textbf{Util.} & \textbf{Rd.}
& \textbf{Suc.} & \textbf{Util.} & \textbf{Rd.}
& \textbf{Suc.} & \textbf{Util.} & \textbf{Rd.}
& \textbf{Suc.} & \textbf{Util.} & \textbf{Rd.} \\
\midrule

Vanilla (LLM)
& 50.0 & $5.0{\scriptstyle\pm15.7}$ & $\mathbf{8.5{\scriptstyle\pm2.4}}$
& 100.0 & $15.0{\scriptstyle\pm35.7}$ & $6.5{\scriptstyle\pm4.8}$
& 100.0 & $35.0{\scriptstyle\pm47.7}$ & $4.5{\scriptstyle\pm2.9}$
& 100.0 & $45.9{\scriptstyle\pm21.5}$ & $\mathbf{2.5{\scriptstyle\pm1.2}}$ \\

Vanilla (SLM)
& 25.0 & $8.8{\scriptstyle\pm20.1}$ & $11.2{\scriptstyle\pm5.9}$
& 75.0 & $37.9{\scriptstyle\pm41.7}$ & $9.9{\scriptstyle\pm7.2}$
& 90.0 & $40.3{\scriptstyle\pm45.0}$ & $4.3{\scriptstyle\pm2.8}$
& 100.0 & $15.0{\scriptstyle\pm30.7}$ & $3.4{\scriptstyle\pm1.5}$ \\

Random
& 85.0 & $40.6{\scriptstyle\pm40.5}$ & $13.9{\scriptstyle\pm9.7}$
& 100.0 & $10.0{\scriptstyle\pm30.0}$ & $4.5{\scriptstyle\pm3.4}$
& 100.0 & $30.0{\scriptstyle\pm45.8}$ & $4.2{\scriptstyle\pm2.2}$
& 100.0 & $43.8{\scriptstyle\pm30.5}$ & $2.5{\scriptstyle\pm1.3}$ \\

\midrule
IQL
& 95.0 & $63.6{\scriptstyle\pm38.1}$ & $10.4{\scriptstyle\pm8.1}$
& 100.0 & $5.0{\scriptstyle\pm21.8}$ & $4.3{\scriptstyle\pm3.5}$
& 100.0 & $0.0{\scriptstyle\pm0.0}$ & $3.1{\scriptstyle\pm1.6}$
& 100.0 & $47.9{\scriptstyle\pm30.5}$ & $2.8{\scriptstyle\pm1.2}$ \\

IQL+SFT
& 100.0 & $69.8{\scriptstyle\pm30.4}$ & $10.1{\scriptstyle\pm7.8}$
& 100.0 & $15.0{\scriptstyle\pm35.7}$ & $7.4{\scriptstyle\pm5.5}$
& 95.0 & $20.0{\scriptstyle\pm40.0}$ & $6.0{\scriptstyle\pm6.5}$
& 100.0 & $51.7{\scriptstyle\pm20.7}$ & $2.5{\scriptstyle\pm1.5}$ \\

IQL+JPO
& 95.0 & $51.7{\scriptstyle\pm34.2}$ & $14.1{\scriptstyle\pm8.1}$
& 90.0 & $\mathbf{40.0{\scriptstyle\pm49.0}}$ & $11.3{\scriptstyle\pm8.7}$
& 100.0 & $35.0{\scriptstyle\pm47.7}$ & $7.4{\scriptstyle\pm6.0}$
& 100.0 & $23.5{\scriptstyle\pm33.9}$ & $5.5{\scriptstyle\pm5.1}$ \\

\textbf{\textsc{EmoDistill} (IQL+SFT+JPO)}
& 90.0 & $\mathbf{72.2{\scriptstyle\pm37.5}}$ & $15.0{\scriptstyle\pm9.8}$
& 100.0 & $30.0{\scriptstyle\pm45.8}$ & $6.5{\scriptstyle\pm4.0}$
& 100.0 & $\mathbf{45.0{\scriptstyle\pm49.7}}$ & $5.5{\scriptstyle\pm3.4}$
& 100.0 & $\mathbf{52.6{\scriptstyle\pm26.6}}$ & $3.1{\scriptstyle\pm2.9}$ \\

\bottomrule
\end{tabular}%
}
\endgroup

\caption{
In-domain negotiation results against the default LLM counterparty.
\textbf{Suc.} denotes Success (\%), \textbf{Util.} denotes Utility (\%), and \textbf{Rd.} denotes negotiation rounds.
Utility is the primary metric and assigns zero to failed negotiations.
Best result per dataset is shown in \textbf{bold}.
}
\label{tab:main_results}
\end{table*}

\begin{table*}[t]
\centering
\setlength{\tabcolsep}{4.2pt}
\renewcommand{\arraystretch}{1.05}
\begingroup
\fontsize{8.2}{9.1}\selectfont
\resizebox{\textwidth}{!}{%
\begin{tabular}{llcccc}
\toprule
\textbf{Block} & \textbf{Method / Condition}
& \textbf{Success (\%)} $\uparrow$
& \textbf{Outcomes (\%)} $\uparrow$
& \textbf{Utility (\%)} $\uparrow$
& \textbf{Rounds} $\downarrow$ \\
\midrule

\multicolumn{6}{l}{\textit{A. Stage-wise training signals}} \\
\midrule

\multirow{3}{*}{IQL}
& Objective trajectory return
& 95.0 & $66.9{\scriptstyle\pm36.0}$ & $\mathbf{63.6{\scriptstyle\pm38.1}}$ & $10.0{\scriptstyle\pm7.2}$ \\
& Episode-judge reward
& 85.0 & $70.6{\scriptstyle\pm27.4}$ & $60.0{\scriptstyle\pm35.7}$ & $10.5{\scriptstyle\pm5.6}$ \\
& Turn-judge reward
& 80.0 & $\mathbf{76.3{\scriptstyle\pm25.4}}$ & $61.0{\scriptstyle\pm38.0}$ & $\mathbf{7.4{\scriptstyle\pm1.9}}$ \\

\midrule
\multirow{3}{*}{IQL+SFT}
& Hybrid quality filter
& 100.0 & $\mathbf{69.8{\scriptstyle\pm30.4}}$ & $\mathbf{69.8{\scriptstyle\pm30.4}}$ & $10.1{\scriptstyle\pm7.8}$ \\
& Episode-judge filter
& 90.0 & $55.1{\scriptstyle\pm33.4}$ & $49.6{\scriptstyle\pm35.7}$ & $9.7{\scriptstyle\pm5.7}$ \\
& Turn-judge filter
& 95.0 & $64.8{\scriptstyle\pm32.0}$ & $61.6{\scriptstyle\pm34.2}$ & $\mathbf{9.0{\scriptstyle\pm3.9}}$ \\

\midrule
\multirow{3}{*}{IQL+SFT+JPO}
& Objective-trajectory advantage
& 85.0 & $64.1{\scriptstyle\pm28.0}$ & $54.5{\scriptstyle\pm34.5}$ & $13.2{\scriptstyle\pm3.4}$ \\
& Episode-judge advantage
& 90.0 & $77.4{\scriptstyle\pm26.0}$ & $69.7{\scriptstyle\pm33.9}$ & $\mathbf{11.2{\scriptstyle\pm3.0}}$ \\
& Turn-judge advantage
& 90.0 & $\mathbf{80.2{\scriptstyle\pm30.3}}$ & $\mathbf{72.2{\scriptstyle\pm37.5}}$ & $15.0{\scriptstyle\pm9.8}$ \\

\midrule
\multicolumn{6}{l}{\textit{B. Emotion-free diagnostic}} \\
\midrule

Vanilla SLM
& No selector / no emotion condition
& 25.0 & $35.3{\scriptstyle\pm26.1}$ & $8.8{\scriptstyle\pm20.1}$ & $11.2{\scriptstyle\pm5.9}$ \\

SFT
& No emotion condition
& 90.0 & $61.1{\scriptstyle\pm28.8}$ & $\mathbf{55.0{\scriptstyle\pm32.5}}$ & $13.7{\scriptstyle\pm10.2}$ \\

JPO
& No emotion condition, no SFT
& 40.0 & $28.7{\scriptstyle\pm30.6}$ & $11.5{\scriptstyle\pm23.4}$ & $24.2{\scriptstyle\pm9.6}$ \\

SFT+JPO
& No emotion condition
& 50.0 & $\mathbf{76.7{\scriptstyle\pm32.2}}$ & $38.4{\scriptstyle\pm46.2}$ & $23.6{\scriptstyle\pm9.1}$ \\

\bottomrule
\end{tabular}%
}
\endgroup

\caption{
Mechanism analysis on CRAD.
\textbf{A:} different stages prefer different training signals: objective trajectory return for IQL, hybrid filtering for SFT, and dense turn-level judge advantages for JPO.
\textbf{B:} removing the explicit emotion selector and emotion condition can preserve high value among successful deals but substantially reduces reliability and Utility after SFT+JPO.
}
\label{tab:mechanism_ablation}
\end{table*}

\begin{table}[t]
\centering
\setlength{\tabcolsep}{5pt}
\renewcommand{\arraystretch}{0.95}

\begingroup
\fontsize{8}{8}\selectfont
\begin{tabular*}{\columnwidth}{@{\extracolsep{\fill}}lccc@{}}
\toprule
\textbf{Method}
& \textbf{$\beta$}
& \textbf{$\Delta U$}
& \textbf{95\% CI} \\
\midrule

DPO  & 0.01  & +18.2 & $[+10.8,+25.4]$ \\
DPO  & 0.001 & +12.1 & $[+4.5,+19.8]$ \\

ORPO & 0.01  & +6.9  & $[-0.5,+14.2]$ \\
ORPO & 0.001 & +11.7 & $[+5.0,+18.5]$ \\

KTO  & 0.01  & +25.3 & $[+17.4,+32.8]$ \\
KTO  & 0.001 & +7.5  & $[+0.1,+15.1]$ \\

\bottomrule
\end{tabular*}

\caption{
Paired Utility gains of \textsc{EmoDistill} over stronger offline-alignment baselines on 100 held-out CRAD scenarios.
}
\label{tab:strong_rl_baselines}
\endgroup
\end{table}

\begin{table*}[t]
\centering
\setlength{\tabcolsep}{5.5pt}
\renewcommand{\arraystretch}{1.05}

\begingroup
\fontsize{8.8}{9.8}\selectfont

\begin{tabular*}{\textwidth}{@{\extracolsep{\fill}}llccccc@{}}
\toprule
\textbf{Counterparty}
& \textbf{Method}
& \textbf{Success (\%)} $\uparrow$
& \textbf{Outcomes (\%)} $\uparrow$
& \textbf{Utility (\%)} $\uparrow$
& \textbf{95\% CI}
& \textbf{Rounds} $\downarrow$ \\
\midrule

\multirow{5}{*}{DeepSeek-V3}
& Vanilla 7B
& 50.0 & 55.6 & 27.8 & $[14.7,42.2]$ & 19.60 \\

& Random 7B
& 95.0 & 34.5 & 32.8 & $[19.8,46.5]$ & 9.75 \\

& IQL 7B
& 70.0 & 27.8 & 19.4 & $[9.6,30.9]$ & 15.00 \\

& IQL+SFT 7B
& 100.0 & 54.3 & 54.3 & $[45.0,62.6]$ & \textbf{7.35} \\

& \textbf{\textsc{EmoDistill}}
& 100.0 & 68.9 & \textbf{68.9} & $\mathbf{[56.4,79.3]}$ & 9.30 \\

\midrule

\multirow{4}{*}{GPT-4o}
& Vanilla 7B
& 70.0 & 50.9 & 35.6 & $[18.2,54.0]$ & 17.35 \\

& IQL 7B
& 100.0 & 44.1 & 44.1 & $[30.3,57.5]$ & \textbf{5.95} \\

& IQL+SFT 7B
& 100.0 & 59.9 & 59.9 & $[50.3,68.5]$ & 7.95 \\

& \textbf{\textsc{EmoDistill}}
& 100.0 & 68.9 & \textbf{68.9} & $\mathbf{[55.5,80.0]}$ & 7.20 \\

\bottomrule
\end{tabular*}

\endgroup

\caption{
Cross-counterparty transfer on \textbf{CRAD} to unseen DeepSeek-V3 and GPT-4o counterparties.
Values are reported on the same $20$ held-out scenarios; 95\% CIs are for Utility.
}
\label{tab:large_llm_counterparty}
\end{table*}

\begin{table*}[t]
\centering
\setlength{\tabcolsep}{2pt}
\setlength{\cmidrulewidth}{0.2pt}
\setlength{\aboverulesep}{1.5pt}
\setlength{\belowrulesep}{1.5pt}
\renewcommand{\arraystretch}{1.05}

\begingroup
\fontsize{7.8}{8}\selectfont
\begin{tabular*}{\textwidth}{@{\extracolsep{\fill}}l|ccc|ccc|ccc|ccc@{}}
\toprule
& \multicolumn{3}{c|}{\textbf{CRAD}} 
& \multicolumn{3}{c|}{\textbf{Disaster}} 
& \multicolumn{3}{c|}{\textbf{Hospital}} 
& \multicolumn{3}{c}{\textbf{Student}} \\
\textbf{Method}
& Suc. & Util. & Rd.
& Suc. & Util. & Rd.
& Suc. & Util. & Rd.
& Suc. & Util. & Rd. \\
\midrule

Vanilla (LLM)
& 50.0  & $5.0{\scriptstyle\pm15.7}$  & $8.5{\scriptstyle\pm2.4}$
& 100.0 & $15.0{\scriptstyle\pm35.7}$ & $6.5{\scriptstyle\pm4.8}$
& 100.0 & $35.0{\scriptstyle\pm47.7}$ & $4.5{\scriptstyle\pm2.9}$
& 100.0 & $45.9{\scriptstyle\pm21.5}$ & $\bm{2.5{\scriptstyle\pm1.2}}$ \\

IQL
& 95.0 & $63.6{\scriptstyle\pm38.1}$ & $10.4{\scriptstyle\pm8.1}$
& 100.0 & $5.0{\scriptstyle\pm21.8}$  & $4.3{\scriptstyle\pm3.5}$
& 100.0 & $0.0{\scriptstyle\pm0.0}$   & $3.1{\scriptstyle\pm1.6}$
& 100.0 & $47.9{\scriptstyle\pm30.5}$ & $2.8{\scriptstyle\pm1.2}$ \\

\textsc{IQL+SFT+JPO}$_\mathrm{C}$
& 90.0  & $\bm{72.2{\scriptstyle\pm37.5}}$ & $15.0{\scriptstyle\pm9.8}$
& 100.0 & $25.0{\scriptstyle\pm43.3}$ & $6.8{\scriptstyle\pm6.2}$
& 100.0 & $25.0{\scriptstyle\pm43.3}$ & $6.5{\scriptstyle\pm6.2}$
& 100.0 & $9.8{\scriptstyle\pm24.1}$  & $3.0{\scriptstyle\pm1.0}$ \\

\textsc{IQL+SFT+JPO}$_\mathrm{D}$
& 95.0  & $57.5{\scriptstyle\pm33.3}$ & $9.9{\scriptstyle\pm6.9}$
& 100.0 & $\bm{30.0{\scriptstyle\pm45.8}}$ & $6.5{\scriptstyle\pm4.0}$
& 100.0 & $25.0{\scriptstyle\pm43.3}$ & $4.8{\scriptstyle\pm2.9}$
& 95.0  & $30.0{\scriptstyle\pm45.8}$ & $6.4{\scriptstyle\pm6.6}$ \\

\textsc{IQL+SFT+JPO}$_\mathrm{H}$
& 100.0 & $64.3{\scriptstyle\pm44.3}$ & $\bm{7.0{\scriptstyle\pm3.5}}$
& 90.0  & $20.0{\scriptstyle\pm40.0}$ & $9.3{\scriptstyle\pm9.5}$
& 100.0 & $\bm{45.0{\scriptstyle\pm49.7}}$ & $5.5{\scriptstyle\pm3.4}$
& 100.0 & $5.0{\scriptstyle\pm21.8}$  & $3.6{\scriptstyle\pm2.1}$ \\

\textsc{IQL+SFT+JPO}$_\mathrm{S}$
& 75.0  & $56.3{\scriptstyle\pm43.6}$ & $14.9{\scriptstyle\pm10.4}$
& 100.0 & $16.8{\scriptstyle\pm28.3}$ & $\bm{2.9{\scriptstyle\pm1.2}}$
& 100.0 & $35.6{\scriptstyle\pm34.7}$ & $\bm{2.9{\scriptstyle\pm1.6}}$
& 100.0 & $\bm{52.6{\scriptstyle\pm26.6}}$ & $3.1{\scriptstyle\pm2.9}$ \\

\bottomrule
\end{tabular*}
\caption{Cross-domain transfer. \textsc{EmoDistill}$_\mathrm{C/D/H/S}$ denotes \textsc{EmoDistill} trained on CRAD, Disaster, Hospital, or Student. Each block reports success, Utility, and rounds; Utility counts failures as 0. Best value per evaluation domain is in \textbf{bold}, except Success.}
\label{tab:cross_domain}
\endgroup
\end{table*}

\begin{table}[t]
\centering
\setlength{\tabcolsep}{3pt}
\renewcommand{\arraystretch}{0.95}

\begingroup
\fontsize{8}{8}\selectfont
\begin{tabular*}{\columnwidth}{@{\extracolsep{\fill}}lccccc@{}}
\toprule
\textbf{Method} & \textbf{$\kappa$} & \textbf{Suc.} & \textbf{Out.} & \textbf{Util.} & \textbf{Rd.} \\
\midrule
JPO-$\kappa$ & 0.00 & 95.0  & $77.0{\scriptstyle\pm24.8}$ & $73.2{\scriptstyle\pm29.7}$ & $9.5{\scriptstyle\pm4.0}$ \\
JPO-$\kappa$ & 0.25 & 100.0 & $69.3{\scriptstyle\pm25.4}$ & $69.3{\scriptstyle\pm25.4}$ & $8.8{\scriptstyle\pm4.3}$ \\
JPO-$\kappa$ & 0.50 & 100.0 & $\mathbf{82.5{\scriptstyle\pm15.9}}$ & $\mathbf{82.5{\scriptstyle\pm15.9}}$ & $9.4{\scriptstyle\pm5.0}$ \\
JPO-$\kappa$ & 0.75 & 95.0  & $80.4{\scriptstyle\pm32.4}$ & $76.4{\scriptstyle\pm36.3}$ & $\mathbf{7.8{\scriptstyle\pm4.3}}$ \\
JPO          & 1.00 & 90.0  & $80.2{\scriptstyle\pm30.3}$ & $72.2{\scriptstyle\pm37.5}$ & $15.0{\scriptstyle\pm9.8}$ \\
\bottomrule
\end{tabular*}
\caption{Risk-controlled JPO on \textbf{CRAD}. }
\label{tab:kappa_risk}
\endgroup
\end{table}

\section{Experimental Results}
\label{sec:results}

We organize the experiments around four questions.
(\textbf{Q1}) Can \textsc{EmoDistill} distill offline LLM-vs-LLM negotiation trajectories into a 7B SLM that improves held-out negotiation Utility over vanilla, component, and stronger offline-alignment baselines?
(\textbf{Q2}) Which parts of the framework drive the gains---state-dependent IQL emotion selection, LoRA-SFT expression distillation, JPO refinement---and which training signal is most effective at each stage?
(\textbf{Q3}) How well do the distilled emotional negotiation skills transfer across domains with different bargaining objectives, preference directions, and concession structures?
(\textbf{Q4}) How robust are the gains to unseen counterparties and alternative reward labelers, and can risk-controlled JPO tune the tradeoff between agreement success and negotiated value?

\paragraph{Q1: In-Domain Negotiation.}
Table~\ref{tab:main_results} and Figure \ref{fig:main_results} show that, among the main component baselines,
\textsc{EmoDistill} achieves the highest Utility on CRAD, Hospital, and Student,
while IQL+JPO performs best on Disaster.
Vanilla and random-emotion agents are less consistent, whereas IQL and SFT provide complementary gains from state-dependent emotion selection and expression learning.
The higher round counts of \textsc{EmoDistill} in some domains indicate a tradeoff between negotiation efficiency and focal-side value.
We further compare against stronger offline-alignment methods on 100 additional held-out CRAD scenarios.
Table~\ref{tab:strong_rl_baselines} shows positive paired Utility gains over all tested DPO, ORPO, and KTO configurations, ranging from $+6.9$ to $+25.3$.
The gains over both DPO settings, ORPO ($\beta=.001$), and KTO ($\beta=.01$) are statistically significant, while the remaining comparisons are positive but inconclusive.
A matched A-LoL refinement is reported in Appendix~\ref{app:alol_baseline}, providing an additional comparison under the same IQL selector and SFT initialization.

\paragraph{Q2: Components and Training Signals.}
Panel A of Table~\ref{tab:mechanism_ablation} shows that supervision is stage-dependent:
objective trajectory return works best for IQL, hybrid filtering for SFT, and turn-level judge advantages for JPO.
This supports separating long-horizon emotion selection from utterance-level expression refinement.
Panel B tests whether explicit emotion contributes beyond generic negotiation distillation.
Removing the emotion condition reduces Utility from $69.8$ to $55.0$ after SFT and from $72.2$ to $38.4$ after SFT+JPO.
Although emotion-free policies can still achieve high Outcomes on successful deals, their lower Success reduces overall Utility.
Thus, explicit emotion acts as a state-dependent control variable that improves the reliability of strategy execution.

\paragraph{Q3: Cross-Domain Transfer.}
Table~\ref{tab:cross_domain} shows that the highest Utility in each domain is obtained by the adapter trained on that domain.
Several off-domain policies retain high Success, but their Utility often decreases substantially.
This suggests partial transfer: general negotiation competence transfers across domains, whereas value extraction remains more sensitive to domain-specific bargaining structure.

\paragraph{Q4: Counterparty, Labeler, and Risk Robustness.}
\label{subsec:robustness}

Table~\ref{tab:large_llm_counterparty} evaluates transfer to unseen
DeepSeek-V3 and GPT-4o counterparties.
\textsc{EmoDistill} reaches $68.9$ Utility against both, retaining the
highest point Utility among the evaluated policies.
We further test whether these gains depend on the counterparty or reward
labeler.
Appendix~\ref{app:judge_counterparty_robustness} reports paired Utility
tests under DeepSeek-V3 and GPT-4o together with a reward-labeler
replacement experiment.
After re-labeling the same 4,000 JPO examples with DeepSeek-V3, the resulting
policy still achieves $100\%$ Success and $63.5$ Utility despite only
$59.3\%$ agreement with the original Qwen preference ordering.
The paired results also show strong gains over Vanilla and IQL, while the
incremental gains over IQL+SFT are smaller and not consistently significant.
However, Utility drops to $13.3$ against Qwen2.5-3B, revealing a smaller-counterparty boundary condition.
Finally, Table~\ref{tab:kappa_risk} shows that risk-controlled JPO provides a tunable success--value tradeoff, with $\kappa=0.5$ yielding the highest CRAD Utility.
Additional analyses are provided in the appendix:
Appendix~\ref{app:stronger_offline_baselines} reports DPO/ORPO/KTO results;
Appendix~\ref{app:alol_baseline} compares A-LoL;
Appendix~\ref{app:judge_counterparty_robustness} tests counterparty and reward-labeler robustness;
Appendix~\ref{app:small_counterparty} reports the Qwen2.5-3B boundary;
Appendices~\ref{app:emotion_free} and \ref{app:emotionfree_covshift} analyze emotion-free behavior;
Appendix~\ref{app:bootstrap_ci} reports bootstrap uncertainty;
Appendix~\ref{app:training_stability} reports training stability;
and Appendices~\ref{app:case_studies} and \ref{app:prompts} provide case studies and prompt details.

\section{Discussion}
\label{sec:discussion}

Our results support treating emotion as an explicit bargaining control variable rather than only a generation style.
IQL selects when to invoke an emotional strategy, while SFT and JPO learn how to realize it in language.
The gains persist across unseen large-model counterparties and a different reward labeler, but transfer remains domain- and counterparty-dependent.
The $\kappa$ analysis further shows that refinement can expose an explicit success--value preference.

\section{Conclusion and Future Work}
\label{sec:conclusion}

We introduced \textsc{EmoDistill}, which distills offline LLM negotiation trajectories into a 7B SLM through IQL emotion selection, LoRA-SFT, and JPO.
Across four domains, it improves Utility over the main baselines in most settings and remains competitive against stronger offline-alignment methods.
Cross-domain and counterparty experiments reveal both transferable negotiation behavior and clear boundary conditions.
Future work should improve transfer to smaller counterparties, reduce dependence on LLM supervision, and evaluate the approach in human-agent negotiation.

\section*{Limitations}
\label{sec:limitations}

Several limitations remain in the current version of \textsc{EmoDistill}.
First, the framework is trained entirely from fixed offline LLM-vs-LLM
trajectories. Although this makes training reusable and avoids costly online
rollouts, the learned selector and expression policy may still encounter
distributional shift at deployment time, especially when the counterparty uses
dialogue strategies not represented in the offline sweep.
Second, \textsc{EmoDistill} relies on an explicit emotion-conditioning channel
at inference time. Our emotion-free ablations show that the adapter can learn
some negotiation behavior without explicit emotion labels, but this behavior is
less reliable. This suggests that emotional negotiation skill is not fully
internalized into the model weights; instead, the explicit emotion variable
remains an important control interface. Developing more robust prompt-free or
latent-emotion variants is an important direction for future work.
Third, transfer is partial rather than universal. Cross-domain experiments show
that success rates often transfer better than value extraction, while utility
remains sensitive to the scalar variable, preference direction, and concession
geometry of each domain. In particular, policies trained on one gap direction
may learn directional anchoring habits that do not automatically flip in domains
with the opposite bargaining geometry. Future work should study domain-adaptive,
sign-aware, or multi-domain training mixtures to improve transfer of value
extraction, not only agreement success.
Fourth, the current evaluation is limited to agent-to-agent negotiation. This is
the intended setting of the paper, but it means that our results should not be
interpreted as evidence that \textsc{EmoDistill} improves human-perceived
negotiation quality. Human studies, multi-judge evaluation, and task-specific
domain expert review would be valuable for understanding how the learned
emotional expressions are perceived outside autonomous agent interactions.
Finally, the framework depends on LLM-judge feedback for dense turn-level
annotation and JPO refinement. While this provides scalable supervision, it also
introduces judge-model dependence and additional cost.

\section*{Ethical Considerations}
\label{sec:ethics}

\textsc{EmoDistill} studies emotion as a strategic control channel in
autonomous agent-to-agent negotiation. This framing has both defensive and
dual-use implications. On the defensive side, modeling emotional influence can
help make user-aligned agents less vulnerable to emotionally framed pressure,
premature concession, or manipulative counterparties. The method is intended to
support agents that preserve their users' stated objectives in adversarial or
high-stakes bargaining settings.
At the same time, a system that learns strategic emotional expression could be
misused to build more persuasive or manipulative negotiation agents. This risk
is especially important in domains involving vulnerable users, financial
decisions, medical access, employment, education, debt, or public services. We
therefore view \textsc{EmoDistill} as appropriate only for bounded
agent-to-agent settings with explicit task objectives, logging, evaluation, and
deployment constraints. It should not be used to manipulate human users, obscure
material information, or pressure people into decisions against their interests.
Our experiments are conducted in synthetic negotiation scenarios between
language-model agents. No real users are negotiated with, and the evaluation
metrics are computed from predefined scenario objectives rather than personal
data.

\bibliography{references}

\clearpage
\appendix

\section{Background}
\label{app:dataset_metrics}

Recent advances in LLM reasoning increasingly move beyond isolated answer generation toward agents that maintain state, retrieve evidence, adapt intermediate decisions, and act over extended interaction trajectories. This trend is reflected in iterative retrieval and reasoning systems~\citep{jiang2024tcragturingcompleteragscasestudy,zhou2026breaksknowledgegraphbased}, agentic reinforcement learning that jointly optimizes reasoning, retrieval, and memory, explicit working-memory mechanisms for long-horizon reasoning~\citep{zhou2026awmanswerableworkingmemory}, and reinforcement-based refinement of inference-time reasoning behavior~\citep{zhou2026refinerunlockallinferencetime}. Complementary work on model attribution and efficient model restructuring further studies how reasoning capability can be identified, selected, and retained under constrained computation~\citep{NEURIPS2025_6724eae9,zhou2025droppingexpertsrecombiningneurons}. Together, these developments suggest that strong agentic reasoning depends not only on the final generated response, but also on controlling intermediate internal and behavioral decisions\citep{zhou2026lookinwardexploreoutward}. In interactive settings, this control extends to the social dimension: an agent must decide when to concede, challenge, reassure, pressure, or coordinate with another agent. Emotion provides a natural interface for such interaction-level reasoning, since different emotional stances can alter both the immediate utterance and the subsequent behavior of the counterpart. We therefore treat emotion not merely as linguistic style, but as a state-dependent social action that can be selected, learned, and distilled into an agent policy. This section defines the notation used by the datasets, policies, and evaluation metrics.

\paragraph{Framework and method names.}
\textsc{EmoDistill} denotes the full offline framework: trajectory collection, judge annotation, offline training, and evaluation. 
In the main tables, \textbf{IQL+SFT+JPO} denotes the full EmoDistill policy. 
Its components are: an IQL emotion selector, a LoRA-SFT expression-policy initializer, and a JPO expression-policy refinement stage.

\paragraph{Negotiation roles.}
The optimized agent is the \emph{focal agent}; the other party is the \emph{counterparty}. 
Domain-specific names such as creditor, debtor, patient, hospital, dispatcher, or student are used only when describing a specific dataset.

\paragraph{Scalar negotiation variable.}
Each scenario contains a scalar negotiation variable $x$, such as overdue days, rescue wait minutes, surgery wait days, or extra hours past 9 PM. 
We denote the counterparty's initial anchor by $x^{\mathrm{opp}}_0$, the focal target by $x^{\star}_{\mathrm{agent}}$, and the final accepted agreement by $x_{\mathrm{final}}$.

\paragraph{Emotion action vocabulary.}
The training sweep uses the full emotion action vocabulary
\[
\mathcal{E}=\{e_1,\ldots,e_{|\mathcal{E}|}\},
\qquad |\mathcal{E}|=28,
\]
consisting of the 28 GoEmotions emotion labels. 
Figure~\ref{fig:emotion_subset} and Appendix~\ref{app:emotion_subset_math} analyze which individual emotions significantly shift CRAD outcomes, but this analysis is not used as a hard filter: all emotions remain available during training, and the IQL selector learns which emotions to upweight or suppress.

\paragraph{Offline sweep.}
For each dataset, the offline sweep is a fixed set of multi-turn trajectories. 
At each focal-agent turn $t$, the sweep stores the dialogue state $s_t$, selected emotion $e_t$, focal-agent utterance $u_t$, counterparty response, terminal outcome, and judge score $r_t$.

\paragraph{Policy levels.}
Emotion-selection policies choose an emotion:
\[
\pi_{\phi}(e_t\mid s_t), \qquad e_t\in\mathcal{E}.
\]
Expression policies generate the utterance:
\[
\pi_{\theta}(u_t\mid s_t,e_t).
\]
The full EmoDistill policy combines these two levels: IQL selects $e_t$, and the SFT-initialized JPO adapter generates $u_t$.

\section{Per-Emotion Prompting Analysis}
\label{app:emotion_subset_math}

This appendix details the controlled prompting study used to support Figure~\ref{fig:emotion_subset}. 
The goal is descriptive: we test whether individual emotion prompts significantly change CRAD negotiation behavior relative to a neutral prompt. 
This analysis is not used to restrict the training action space; \textsc{EmoDistill} uses the full $|\mathcal{E}|=28$ vocabulary, consisting of the 28 GoEmotions emotion labels.

For each emotion $e$ and each CRAD test scenario $s\in\{1,\ldots,20\}$, we run $20$ sampled negotiations. 
Let $r_{e,s,j}$ denote the normalized judge reward or utility from run $j$ under emotion $e$ on scenario $s$. 
We first compute the per-scenario mean
\begin{equation}
\bar r_{e,s}=\frac{1}{20}\sum_{j=1}^{20} r_{e,s,j},
\end{equation}
and then the overall emotion mean
\begin{equation}
\hat\mu_e=\frac{1}{20}\sum_{s=1}^{20}\bar r_{e,s}.
\end{equation}
Figure~\ref{fig:emotion_subset} ranks emotions by $\hat\mu_e$ and plots $95\%$ confidence intervals over scenario-level means.

To test whether an emotion differs from neutral, we use paired scenario-level differences:
\begin{equation}
\delta_{e,s}=\bar r_{e,s}-\bar r_{\mathrm{neutral},s}.
\end{equation}
We then apply a paired $t$-test over the $20$ scenarios:
\begin{equation}
t_e=\frac{\bar\delta_e}{s_{\delta_e}/\sqrt{20}},
\qquad
\bar\delta_e=\frac{1}{20}\sum_s \delta_{e,s}.
\end{equation}
Because we test the 27 non-neutral emotions against the neutral baseline, we apply Bonferroni correction with threshold
\begin{equation}
\alpha_{\mathrm{Bonf}}=\frac{0.05}{27}\approx 0.00185.
\end{equation}

This analysis identifies multiple emotions whose effects are significantly above the neutral baseline, confirming that emotional framing can systematically shift negotiation outcomes. 
We use this result only as motivation for emotion-conditioned policy learning. 
During training, all $28$ emotions remain in the action vocabulary, and the IQL selector learns which emotions to upweight or suppress from the offline sweep.

\section{Detailed Algorithm}
\label{app:compared_policies}

This section gives a uniform mathematical specification for every method evaluated in the main results. 
Notation follows Appendix~\ref{app:dataset_metrics}: $s_t$ is the dialogue state at focal-agent turn $t$, $e_t\in\mathcal{E}$ ($|\mathcal{E}|{=}28$) is the emotion token, $u_t$ is the focal-agent utterance, $r_t$ is the per-turn metric-aligned judge score, $A_t=(r_t-\mu_{\mathrm{scen}})/(\sigma_{\mathrm{scen}}+\epsilon)$ is its scenario-wise z-score, and $R(\tau)$ is the outcome-shaped trajectory return defined in Appendix~\ref{app:reward_design}. 
The evaluation metrics derived from the final agreement are defined separately in Appendix~\ref{app:normalized_savings}. 
$\pi_{\mathrm{LLM}}(u_t\mid s_t,e_t)$ denotes the frozen base utterance policy used during sweep construction; $\pi_\theta(u_t\mid s_t,e_t)$ denotes the LoRA-augmented Qwen2.5-7B expression policy with $\theta$ the LoRA parameters. 
Emotion selectors are written $\pi_\phi(e_t\mid s_t)$.

\paragraph{Prompt-only controls.}
Neutral prompting uses a fixed neutral emotion at every focal-agent turn. 
Random-emotion prompting samples uniformly from $\mathcal{E}$. 
These are calibration controls rather than offline-learning methods:
\begin{equation}
\label{eq:controls}
\begin{aligned}
\pi_{\mathrm{neu}}(u_t\mid s_t)
&=\pi_{\mathrm{LLM}}(u_t\mid s_t,\,e_t{=}\text{neutral}),\\
\pi_{\mathrm{rnd}}(u_t\mid s_t)
&=\mathbb{E}_{e_t\sim\mathrm{Unif}(\mathcal{E})}\!\big[\pi_{\mathrm{LLM}}(u_t\mid s_t,e_t)\big].
\end{aligned}
\end{equation}
No parameters are learned.

\paragraph{IQL selector.}
IQL learns an offline emotion-selection policy over state--emotion pairs. 
It chooses the emotion to inject at each turn but does not update the utterance generator. 
For a trajectory $\tau$, the IQL reward is terminal:
\begin{equation}
\label{eq:iql_terminal_app}
\bar r^{\mathrm{IQL}}_t =
\begin{cases}
R(\tau), & t=T_\tau,\\
0, & \text{otherwise},
\end{cases}
\end{equation}
where $T_\tau$ is the final focal-agent turn. 
With expectile parameter $\tau_{\mathrm{exp}}$, IQL optimizes
\begin{equation}
\label{eq:iql_vq}
\begin{aligned}
\mathcal{L}_V(\psi)
&= \mathbb{E}_{\mathcal{D}}\!\left[
L_2^{\tau_{\mathrm{exp}}}\!\left(Q_{\bar\theta}(s,e)-V_\psi(s)\right)
\right],\\
\mathcal{L}_Q(\theta)
&= \mathbb{E}_{\mathcal{D}}\!\left[
\left(\bar r^{\mathrm{IQL}}_t+\gamma V_\psi(s')-Q_\theta(s,e)\right)^2
\right],
\end{aligned}
\end{equation}
where $L_2^{\tau_{\mathrm{exp}}}(x)=|\tau_{\mathrm{exp}}-\mathbb{1}\{x<0\}|x^2$. 
The selector is extracted by advantage-weighted regression (AWR) with temperature $\beta_{\mathrm{IQL}}$:
\begin{equation}
\label{eq:iql_pi}
\begin{aligned}
\pi^{\mathrm{IQL}}_\phi(e\mid s)
&\propto \exp\!\left(\beta_{\mathrm{IQL}} A(s,e)\right),\\
A(s,e)
&= Q_\theta(s,e)-V_\psi(s).
\end{aligned}
\end{equation}
At inference, $e_t\sim\pi^{\mathrm{IQL}}_\phi(\cdot\mid s_t)$ and $u_t\sim\pi_{\mathrm{LLM}}(\cdot\mid s_t,e_t)$. 
The LoRA expression adapter is not used in the IQL-only ablation.

\paragraph{LoRA-SFT expression policy.}
SFT trains the expression policy by supervised learning on high-quality turn-level demonstrations. 
Let $\mathcal{D}_{\mathrm{top}}\subset\mathcal{D}$ denote the top-$25\%$ of $(s_t,e_t,u_t)$ tuples ranked by the hybrid filtering score $q_t^{\mathrm{hyb}}$ in Appendix~\ref{app:sft_filter_signal}. 
The SFT objective is token-level cross-entropy under emotion conditioning:
\begin{equation}
\label{eq:sft_app}
\begin{aligned}
\mathcal{L}_{\mathrm{SFT}}(\theta)
&= -\,\mathbb{E}_{(s_t,e_t,u_t)\sim\mathcal{D}_{\mathrm{top}}}
\Bigg[
\sum_{k=1}^{|u_t|} \\
&\qquad
\log \pi_\theta\!\left(
u_{t,k}\mid s_t,e_t,u_{t,<k}
\right)
\Bigg].
\end{aligned}
\end{equation}
At inference, $e_t\sim\pi^{\mathrm{IQL}}_\phi(\cdot\mid s_t)$, then $u_t\sim\pi_\theta(\cdot\mid s_t,e_t)$.

\paragraph{\textsc{EmoDistill}.}
\textsc{EmoDistill} is our main method. 
It uses IQL for emotion selection and two-stage LoRA training for expression-policy optimization. 
Stage~1 is the SFT objective above. 
Stage~2 freezes the SFT snapshot $\pi_{\mathrm{ref}}=\pi^{\mathrm{SFT}}_\theta$ and applies a PPO-clipped surrogate with a K3 KL anchor:
\begin{equation}
\label{eq:rho_app}
\rho_t(\theta)=\frac{\pi_\theta(u_t\mid s_t,e_t)}{\pi_{\mathrm{ref}}(u_t\mid s_t,e_t)}.
\end{equation}
JPO uses the asymmetric advantage
\begin{equation}
\label{eq:asym_adv_app}
\widetilde A_t =
\begin{cases}
A_t, & A_t>0,\\
\kappa A_t, & A_t\le 0,
\end{cases}
\end{equation}
where $\kappa\in[0,1]$ controls the weight on negative-advantage samples. 
The objective is
\begin{equation}
\label{eq:jpo_app}
\begin{aligned}
\mathcal{L}_{\mathrm{JPO}}(\theta)
=
&-\,\mathbb{E}_{\mathcal{D}_{\mathrm{top}}}\!\Big[
\min\!\big(
\rho_t \widetilde A_t,\;
\mathrm{clip}(\rho_t,1{-}\epsilon,1{+}\\\epsilon)\widetilde A_t
\big)
\Big]
&+\,\lambda_{\mathrm{KL}}\cdot
\mathbb{E}\!\big[\mathrm{KL}_{\mathrm{K3}}(\pi_\theta\,\|\,\pi_{\mathrm{ref}})\big].
\end{aligned}
\end{equation}
Inference: $e_t\sim\pi^{\mathrm{IQL}}_\phi(\cdot\mid s_t)$, then $u_t\sim\pi_\theta(\cdot\mid s_t,e_t)$.

\paragraph{IQL+JPO ablation.}
The \textbf{IQL+JPO} ablation removes the SFT warm start and tests direct judge-guided refinement. 
The same clipped objective is used, but the reference policy is the base instruction model rather than the SFT adapter. 
This isolates whether dense judge advantages are sufficient without imitation-based initialization.

\paragraph{Emotion-Free EmoDistill.}
Emotion-Free EmoDistill removes the explicit emotion block and the inference-time selector. 
It is a diagnostic ablation that tests internalization of emotional strategy. 
The same two-stage objective as \textsc{EmoDistill} is applied to states with the emotion block stripped:
\begin{equation}
\label{eq:promptfree_sft_app}
\begin{aligned}
\mathcal{L}^{\mathrm{free}}_{\mathrm{SFT}}(\theta)
&= -\,\mathbb{E}_{(s_t,e_t,u_t)\sim\mathcal{D}_{\mathrm{top}}}
\Bigg[
\sum_k \\
&\qquad
\log \pi_\theta\!\left(
u_{t,k}\mid s_t,u_{t,<k}
\right)
\Bigg].
\end{aligned}
\end{equation}
The free-form JPO ratio is
\begin{equation}
\label{eq:promptfree_rho_app}
\rho_t^{\mathrm{free}}(\theta)
= \frac{
\pi_\theta(u_t\mid s_t)
}{
\pi^{\mathrm{free}}_{\mathrm{ref}}(u_t\mid s_t)
}.
\end{equation}
The corresponding JPO objective is
\begin{equation}
\label{eq:promptfree_jpo_app}
\begin{aligned}
\mathcal{L}^{\mathrm{free}}_{\mathrm{JPO}}(\theta)
&= -\,\mathbb{E}\!\Bigg[
\min\!\Big(
\rho_t^{\mathrm{free}}\widetilde A_t,\; \\
&\qquad
\operatorname{clip}\!\left(
\rho_t^{\mathrm{free}},1{-}\epsilon,1{+}\epsilon
\right)\widetilde A_t
\Big)
\Bigg] \\
&\quad
+\,\lambda_{\mathrm{KL}}\,
\mathrm{KL}_{\mathrm{K3}}\!\left(
\pi_\theta\,\|\,\pi^{\mathrm{free}}_{\mathrm{ref}}
\right).
\end{aligned}
\end{equation}
At inference, no $e_t$ is sampled and no selector is consulted: $u_t\sim\pi_\theta(\cdot\mid s_t)$. 
The covariate-shift consequences of this design are formalized in Appendix~\ref{app:emotionfree_covshift}.

\begin{table*}[h]
\centering
\small
\setlength{\tabcolsep}{4pt}
\begin{tabular}{lccc}
\toprule
\textbf{Method} & \textbf{Updates} & \textbf{Training signal} & \textbf{Inference cond.} \\
\midrule
Vanilla    & ---            & ---                              & $(s_t,\,e_t{=}\text{neu})$ \\
Random emotion    & ---            & ---                              & $(s_t,\,e_t{\sim}U(\mathcal{E}))$ \\
IQL selector      & \textsc{Emo}   & terminal $R(\tau)$               & $(s_t,\,e_t{\sim}\pi^{\mathrm{IQL}}_\phi)$ \\
IQL+SFT           & \textsc{Emo}+\textsc{LoRA}  & top-$25\%$ hybrid BC       & $(s_t,\,e_t{\sim}\pi^{\mathrm{IQL}}_\phi)$ \\
IQL+JPO           & \textsc{Emo}+\textsc{LoRA}  & per-turn $\widetilde A_t$  & $(s_t,\,e_t{\sim}\pi^{\mathrm{IQL}}_\phi)$ \\
\textsc{EmoDistill}  & \textsc{Emo}+\textsc{LoRA}  & top-$25\%$ hybrid BC $+$ per-turn $\widetilde A_t$ & $(s_t,\,e_t{\sim}\pi^{\mathrm{IQL}}_\phi)$ \\
Emotion-Free EmoDistill & \textsc{LoRA}  & top-$25\%$ hybrid BC $+$ per-turn $\widetilde A_t$ & $s_t$ only \\
\bottomrule
\end{tabular}
\caption{
What each compared policy learns.
\textsc{Emo} denotes the emotion selector and
\textsc{LoRA} denotes the trainable utterance adapter.
``Inference cond.'' specifies the state and emotion information
available to the utterance policy at deployment.
}
\label{tab:learnable_summary}
\end{table*}

\subsection{Algorithmic Summary}
\label{app:algorithms}

\begin{algorithm}[h]
\caption{IQL Emotion Selector}
\label{alg:iql}
\begin{algorithmic}[1]
\Require{Offline sweep $\mathcal{D}$, outcome-shaped returns $R(\tau)$, expectile $\tau_{\mathrm{exp}}$, AWR temperature $\beta$}
\For{$N$ gradient steps}
    \State Update $V_{\psi}$ by expectile regression.
    \State Update $Q_{\theta}$ using terminal-reward TD targets.
    \State Compute $A(s,e)=Q_{\theta}(s,e)-V_{\psi}(s)$.
    \State Update $\pi_{\phi}$ by advantage-weighted regression.
\EndFor
\State \Return emotion selector $\pi_{\phi}(e\mid s)$
\end{algorithmic}
\end{algorithm}

\begin{algorithm}[h]
\caption{SFT Expression Policy}
\label{alg:sft}
\begin{algorithmic}[1]
\Require{Offline sweep $\mathcal{D}$, hybrid scores $q_t^{\mathrm{hyb}}$, base SLM $\pi_0$}
\State Rank $(s_t,e_t,u_t)$ tuples by $q_t^{\mathrm{hyb}}$.
\State Retain the top $25\%$ as $\mathcal{D}_{\mathrm{top}}$.
\State Initialize LoRA adapter $\theta$ on $\pi_0$.
\For{$N$ supervised steps}
    \State Sample $(s_t,e_t,u_t)$ from $\mathcal{D}_{\mathrm{top}}$.
    \State Minimize token-level cross-entropy on $u_t$.
\EndFor
\State \Return SFT adapter $\pi_{\mathrm{SFT}}$
\end{algorithmic}
\end{algorithm}

\begin{algorithm}[h]
\caption{Judge Policy Optimization}
\label{alg:jpo}
\begin{algorithmic}[1]
\Require{Sweep tuples $\{(s_t,e_t,u_t,A_t)\}$, reference policy $\pi_{\mathrm{ref}}$, clip $\epsilon$, KL coefficient $\lambda_{\mathrm{KL}}$, negative-advantage coefficient $\kappa$}
\State Initialize $\pi_{\theta}$ from the SFT adapter.
\State Freeze $\pi_{\mathrm{ref}}=\pi_{\mathrm{SFT}}$.
\For{$N$ gradient steps}
    \State Sample minibatch $\{(s_t,e_t,u_t,A_t)\}\sim\mathcal{D}_{\mathrm{top}}$.
    \State Compute $\widetilde A_t=A_t$ if $A_t>0$, else $\widetilde A_t=\kappa A_t$.
    \State Compute $\rho_t=\pi_{\theta}(u_t\mid s_t,e_t)/\pi_{\mathrm{ref}}(u_t\mid s_t,e_t)$.
    \State Compute clipped policy loss:
    \[
    \mathcal{L}_{\mathrm{clip}}
    =
    -
    \min
    \left(
    \rho_t \widetilde A_t,
    \operatorname{clip}(\rho_t,1-\epsilon,1+\epsilon)\widetilde A_t
    \right).
    \]
    \State Compute K3 KL anchor:
    \[
    \mathcal{L}_{\mathrm{KL}}
    =
    \mathrm{KL}_{\mathrm{K3}}
    \left(
    \pi_{\theta}\,\|\,\pi_{\mathrm{ref}}
    \right).
    \]
    \State Update $\theta$ using
    \[
    \mathcal{L}_{\mathrm{JPO}}
    =
    \mathcal{L}_{\mathrm{clip}}
    +
    \lambda_{\mathrm{KL}}\mathcal{L}_{\mathrm{KL}}.
    \]
\EndFor
\State \Return JPO adapter $\pi_{\theta}$
\end{algorithmic}
\end{algorithm}

\section{Reward Design and Training Signals}
\label{app:reward_design}

The same offline sweep provides different signals for different stages. 
For IQL, a signal is used as an emotion-selection reward; for SFT, it is used as a data-filtering score; for JPO, it becomes an utterance-level advantage. 
This section gives the full signal definitions behind the training-signal ablation in Panel A of Table~\ref{tab:mechanism_ablation}. 
The per-turn judge prompt that produces $r_t$ is shown verbatim in Appendix~\ref{app:prompts_judge}.

\subsection{Emotion-Selection Rewards}
\label{app:terminal_reward}

Offline value-based selectors such as IQL attach rewards to selected emotions at each state. 
We compare three signal placements. 
The \emph{outcome-shaped} variant uses the objective trajectory return $R(\tau)$ only at the terminal focal-agent turn and propagates it through Bellman backups. 
The \emph{episode-judge} variant broadcasts one dialogue-level judge score to every focal turn. 
The \emph{turn-judge} variant uses the judge score assigned to the current focal utterance. 
The default IQL selector uses the outcome-shaped reward so that emotion selection is tied to actual bargaining movement rather than surface persuasiveness alone.

\subsection{SFT Filtering Signal}
\label{app:sft_filter_signal}

SFT does not use the signal as a reinforcement-learning reward. 
Instead, it uses a hybrid quality score to select high-quality turn-level demonstrations:
\begin{equation}
\label{eq:sft_filter_hybrid}
q^{\mathrm{hyb}}_t = r_t + \tfrac{1}{2}R(\tau).
\end{equation}
Here $r_t$ measures the local quality of the focal utterance under the metric-aligned judge, while $R(\tau)$ measures whether the trajectory as a whole produces favorable bargaining dynamics and terminal agreement. 
We retain the top $25\%$ of $(s_t,e_t,u_t)$ tuples ranked by $q_t^{\mathrm{hyb}}$. 
This filter avoids imitating utterances that sound locally persuasive but occur inside globally unproductive negotiations, while also avoiding purely terminal filtering that would keep weak individual turns from successful trajectories.

\subsection{JPO Advantage Signals}
\label{app:jpo_reward_variants}

For JPO, the signal is used as an offline policy-improvement advantage. 
We compare three variants: outcome-shaped advantage, episode-judge advantage, and turn-judge advantage.

\paragraph{Outcome-shaped advantage.}
The outcome-shaped reward is a sum of time-weighted per-step shaping plus a non-time-weighted terminal anchor:
\begin{equation}
\label{eq:outcome_shaped_reward}
R(\tau)
=
\underbrace{
\sum_{t=1}^{T_\tau} w(t)
\bigl(
\Delta^{\mathrm{ctp}}_t-\Delta^{\mathrm{foc}}_t
\bigr)
}_{\text{time-weighted step shaping}}
+
R^{\mathrm{term}}(\tau).
\end{equation}
Let $g=x^{\star}_{\mathrm{agent}}-x^{\mathrm{opp}}_0$ be the signed anchor-to-target gap and $d=\max(1,|g|)$ its stabilized magnitude. 
The sign of $g$ makes the formulation valid whether the focal agent wants a smaller or larger scalar value. 
The normalized counterparty concession is
\begin{equation}
\label{eq:counterparty_step}
\Delta^{\mathrm{ctp}}_t
=
\mathrm{clip}\!\left(
\frac{\operatorname{sgn}(g)(x^{\mathrm{ctp}}_t-x^{\mathrm{ctp}}_{t-1})}{d},
-2,2
\right),
\end{equation}
which is positive when the counterparty moves toward the focal target. 
The normalized focal retreat is
\begin{equation}
\label{eq:focal_step}
\Delta^{\mathrm{foc}}_t
=
\mathrm{clip}\!\left(
\frac{-\operatorname{sgn}(g)(x^{\mathrm{foc}}_t-x^{\mathrm{foc}}_{t-1})}{d},
-2,2
\right),
\end{equation}
which is positive when the focal agent moves away from its own target. 
Thus, $\Delta^{\mathrm{ctp}}_t-\Delta^{\mathrm{foc}}_t$ rewards turns where the counterparty concedes more than the focal agent retreats.

The implicit dialogue-length penalty is the \emph{linear time-decay weight}
\begin{equation}
\label{eq:length_penalty_weight}
\begin{aligned}
w(t)
&= \max\!\left(0,\min\!\left(1,1-\frac{t}{T_{\max}}\right)\right),\\
T_{\max} &= 30.
\end{aligned}
\end{equation}
Late concessions therefore earn less step credit, providing a length penalty without an explicit additive term. 
The terminal anchor is held constant so late but successful closes are not double-penalized:
\begin{equation}
\label{eq:terminal_anchor}
R^{\mathrm{term}}(\tau)
=
\begin{cases}
+2, & \text{if agreement reached,}\\
-2, & \text{otherwise.}
\end{cases}
\end{equation}
Importantly, $R(\tau)$ uses no LLM-judge signal.

\paragraph{Episode-judge advantage.}
A metric-aligned LLM judge scores the whole dialogue, and the resulting dialogue-level score is broadcast to all focal-agent turns. 
This signal is subjective and judge-based, but it does not provide turn-level credit assignment.

\paragraph{Turn-judge advantage.}
The judge scores each focal-agent turn independently under the metric-aligned rubric in Appendix~\ref{app:prompts_judge}. 
The rubric rewards anchoring near the focal target, concrete proposals, and scenario-grounded leverage, and penalizes capitulation, repetition, vagueness, and emotionally inappropriate concessions. 
Scores are normalized within each scenario:
\begin{equation}
\label{eq:advantage_app}
A_t
=
\frac{r_t-\mu_{\mathrm{scenario}}}{\sigma_{\mathrm{scenario}}+\epsilon}.
\end{equation}
The normalized score $A_t$ is used as the fixed offline advantage in JPO, and its asymmetric form $\widetilde A_t$ is defined in Eq.~\eqref{eq:asym_adv_app}.

\section{Experimental Setup}
\label{app:experimental_setup}

\subsection{Datasets}
\label{app:dataset_details}

We evaluate on four datasets from CRAD and EmoMAS: Credit Recovery, Disaster Rescue, Hospital Surgery Scheduling, and Student Sleep Scheduling. 
Each dataset contains $100$ scenarios split into $80$ training scenarios and $20$ held-out test scenarios.

\begin{table*}[h]
\centering
\small
\begin{tabular}{lccc}
\toprule
\textbf{Dataset} & \textbf{N} & \textbf{Gap sign} & \textbf{Quantity} \\
\midrule
Credit Recovery (CRAD)        & 100 & Target $<$ Anchor & overdue days \\
Disaster Rescue               & 100 & Target $>$ Anchor & wait minutes \\
Hospital Surgery Scheduling   & 100 & Target $>$ Anchor & wait days \\
Student Sleep Scheduling      & 100 & Target $<$ Anchor & extra hours past 9\,PM \\
\bottomrule
\end{tabular}
\caption{Datasets used in our experiments. The \textsc{Gap sign} column indicates whether the focal agent's target is lower or higher than the counterparty's initial anchor, i.e., the sign of $x^{\star}_{\mathrm{agent}}-x^{\mathrm{opp}}_0$.}
\label{tab:datasets}
\end{table*}

The sign of the anchor-to-target gap differs across domains. 
In Credit Recovery and Student Sleep Scheduling, the focal agent prefers a smaller value than the counterparty's initial anchor, so $x^{\star}_{\mathrm{agent}}-x^{\mathrm{opp}}_0<0$. 
In Disaster Rescue and Hospital Surgery Scheduling, the focal agent prefers a larger value, so $x^{\star}_{\mathrm{agent}}-x^{\mathrm{opp}}_0>0$. 
The normalized savings formula in Appendix~\ref{app:normalized_savings} is sign-invariant by construction and remains valid in both regimes.

\subsection{Offline Sweep Construction}
\label{app:sweep_construction}

For each dataset, we run $80$ training scenarios $\times$ $100$ emotion-sequence seeds. 
Each seed fixes a sampled sequence from the full $|\mathcal{E}|=28$ vocabulary, consisting of the 28 GoEmotions emotion labels. 
Thus, each domain contains $8000$ offline trajectories. 
The $100$ factor refers to random emotion-sequence seeds per training scenario, not model-training seeds.

Each rollout records:
\begin{itemize}[leftmargin=*]
    \item scenario identifier and dataset;
    \item focal-agent target $x^{\star}_{\mathrm{agent}}$;
    \item counterparty initial anchor $x^{\mathrm{opp}}_0$;
    \item dialogue history at each focal-agent turn;
    \item selected emotion $e_t$;
    \item focal-agent utterance $u_t$;
    \item counterparty response;
    \item terminal outcome, including final agreement, success, and rounds;
    \item per-turn LLM-judge score $r_t$.
\end{itemize}

All sweep-generation calls use Qwen3.5-Plus through DashScope. 
We use a $6$-key round-robin setup to reduce rate-limit bottlenecks. 
Each negotiation is capped at $30$ turns. 
The same sweep is reused for IQL, SFT, and JPO.

\subsection{Evaluation Metrics}
\label{app:evaluation_protocol}

Each method is evaluated on the same $20$ held-out scenarios per dataset. 
We report success rate, \textbf{Outcomes}, \textbf{Utility}, and mean dialogue rounds.

\subsection{Success Rate}

A mediator classifies each dialogue as \textit{accepted}, \textit{breakdown}, or \textit{ongoing}. 
Success rate is the fraction of dialogues that reach a valid accepted agreement satisfying the task-specific criterion. 
Dialogues classified as \textit{breakdown} or \textit{ongoing} are counted as unsuccessful.

\subsection{Normalized Savings}
\label{app:normalized_savings}

Let $x^{\mathrm{opp}}_0$ be the counterparty's initial anchor, $x^{\star}_{\mathrm{agent}}$ be the focal target, and $x_{\mathrm{final}}$ be the final accepted agreement. 
For successful negotiations, normalized savings is the fraction of the anchor-to-target distance closed by the final agreement:
\begin{equation}
\label{eq:sav}
\textsc{Sav}
=
\frac{x_{\mathrm{final}}-x^{\mathrm{opp}}_0}
{x^{\star}_{\mathrm{agent}}-x^{\mathrm{opp}}_0}.
\end{equation}
The metric is sign-invariant: if the focal target is lower than the anchor, both numerator and denominator are negative for progress toward the target; if the focal target is higher, both are positive. 
Equivalently,
\begin{equation}
\label{eq:sav_signinv}
\textsc{Sav}
=
\frac{
\left|x_{\mathrm{final}}-x^{\mathrm{opp}}_0\right|
}{
\left|x^{\star}_{\mathrm{agent}}-x^{\mathrm{opp}}_0\right|
}
\end{equation}
whenever the final agreement lies between the anchor and target. 
We clip values outside $[0,1]$ for aggregation.

\subsection{Outcomes and Utility}

\textbf{Outcomes} averages $\textsc{Sav}$ over successful negotiations only. 
\textbf{Utility} averages over all $20$ held-out scenarios and assigns zero utility to failed negotiations:
\begin{equation}
\label{eq:utility_per_episode}
u_i =
\begin{cases}
\textsc{Sav}_i, & \text{if episode } i \text{ succeeds},\\
0, & \text{otherwise}.
\end{cases}
\end{equation}
Utility is the stricter main metric because it captures both agreement quality and failure risk.

\subsection{Mean Dialogue Rounds}

Mean dialogue rounds measures negotiation efficiency. 
It should be interpreted together with Utility, since a policy can reduce rounds by accepting too early.

\section{Model, Training Setup, and Hyperparameters}
\label{app:model_training_setup}

\paragraph{Backbones.}
Qwen3.5-Plus is used for construction of the default offline sweep,
default reward annotation, and the default in-domain counterparty.
The trainable expression policy is Qwen2.5-7B-Instruct with LoRA
adapters.
Cross-counterparty evaluations additionally use DeepSeek-V3,
GPT-4o, and Qwen2.5-3B-Instruct.
DeepSeek-V3 is also used as an alternative reward labeler in the
robustness experiment of
Appendix~\ref{app:judge_counterparty_robustness}.

\paragraph{LoRA configuration.}
We apply LoRA adapters to \texttt{q\_proj}, \texttt{k\_proj}, \texttt{v\_proj}, and \texttt{o\_proj}. 
Unless otherwise stated, we use rank $16$. 
We sweep ranks $\{4,16,64\}$ and find rank $16$ gives the best validation tradeoff.

\paragraph{SFT.}
The SFT ablation behavior-clones high-quality focal-agent utterances from the offline sweep. 
It is implemented with LoRA, but we refer to the method as SFT because LoRA is the parameter-efficient fine-tuning mechanism.

\paragraph{JPO.}
JPO starts from the SFT adapter and optimizes the PPO-clipped offline
objective from Section~\ref{sec:expression}.
The reference policy is the frozen SFT adapter.
We use clipping parameter $\varepsilon=0.2$.
We sweep
$\lambda_{\mathrm{KL}}\in\{0.01,0.04,0.1,0.5\}$
and use $\lambda_{\mathrm{KL}}=0.04$ for the default model.
For the asymmetric advantage in Eq.~\eqref{eq:asym_adv_app},
the canonical JPO model is trained with $\kappa=1$.
The risk-control ablation trains variants with
$\kappa\in\{0,0.25,0.5,0.75,1\}$, as reported in
Table~\ref{tab:kappa_risk}.
The coefficient $\kappa$ affects refinement training only and is not
an inference-time parameter.

\paragraph{IQL.}
We tune expectile $\tau\in\{0.7,0.8,0.9\}$, AWR temperature $\beta\in\{1,3,10\}$, and discount $\gamma\in\{0.95,0.99\}$.



\section{Why Offline RL}
\label{app:why_offline}

Online policy optimization would require fresh multi-turn negotiations and fresh LLM-judge calls at every gradient step. Offline JPO avoids this by reusing pre-judged tuples from the fixed sweep. This section provides the cost decomposition and explains why we do not run pure online PPO, GRPO, or DAgger-style distillation.

\paragraph{Offline JPO cost.}
A single offline JPO step samples pre-judged tuples $(s_t,e_t,u_t,A_t)$, forwards them through Qwen2.5-7B with LoRA, computes the clipped surrogate and KL anchor, and backpropagates. On a single A100, this costs approximately $5$ seconds per optimizer step. The expensive LLM interactions and judge annotations are paid once during sweep construction and then reused across SFT, JPO and IQL.

\paragraph{Pure on-policy PPO or GRPO.}
A pure on-policy variant would require new negotiations at every gradient step. For batch size $b=16$, each step would need $16$ fresh negotiations with the current focal-agent policy, counterparty calls, judge calls for every turn, reward normalization, and then the same gradient update. In our sweep, a negotiation takes approximately $75$ seconds wall-clock, and judge scoring adds approximately $10$ seconds per turn for an average of $18$ turns. With concurrency $c=6$, the approximate per-step cost is
\begin{equation}
\label{eq:onpol_cost}
T_{\mathrm{onpol}}^{\mathrm{step}}
\;\approx\;
\frac{16\times(75+180)}{6}
\;\approx\;
680\,\mathrm{s}.
\end{equation}
This is roughly $130\times$ slower than offline JPO. For a $2000$-step run, this would become roughly $15$ GPU-days of API-limited wall-clock time per dataset before rate limits become the bottleneck. GRPO removes the value network but does not remove the need for fresh group rollouts and judge calls, so its wall-clock bottleneck is similar in our setting.

\paragraph{Stability considerations.}
Pure online optimization would also be less stable because the model would optimize directly against a noisy $1$--$10$ judge signal without the strong SFT initialization and KL anchor used by JPO. Offline JPO is appropriate here because the per-turn judge score can be computed once for each observed tuple and reused across optimization steps. The main risk is covariate shift between the teacher-induced sweep distribution and the student-induced deployment distribution. We mitigate this risk with SFT initialization, a tight KL anchor, and prompt-conditional deployment; Prompt-Free EmoDistill intentionally removes this conditioning.

\section{Emotion-Free EmoDistill and Covariate Shift}
\label{app:emotionfree_covshift}

Emotion-conditional EmoDistill keeps the IQL selector at inference: the selector chooses an emotion, the prompt includes the corresponding emotional approach, and the JPO-trained expression adapter generates the focal-agent utterance. 
Emotion-free EmoDistill removes both the emotion block and the inference-time selector. 
It tests whether the LoRA adapter can internalize emotional strategy without explicit emotion conditioning.
The source of covariate shift is that the offline utterances were generated under emotion conditioning,
\[
u_t \sim \pi_{\mathrm{LLM}}(u_t\mid s_t,e_t),
\]
while emotion-free training and inference condition only on the stripped state,
\[
u_t \sim \pi_{\theta}^{\mathrm{free}}(u_t\mid s_t).
\]
Thus, the student must imitate outputs whose causal emotional condition is hidden. 
Equivalently, the emotion-free target is an emotion-marginalized teacher:
\[
\pi_{\beta}(u_t\mid s_t)
=
\sum_{e_t\in\mathcal{E}}
\pi_{\mathrm{emo}}(e_t\mid s_t)
\pi_{\mathrm{LLM}}(u_t\mid s_t,e_t).
\]
If multiple emotions produce distinct high-quality utterances for similar states, the emotion-free student is forced to average over hidden modes.

This multimodality is strong in practice. 
On CRAD, replaying recorded focal-agent turns through the IQL selector shows that only $136/2104$ turns ($6.5\%$) match the greedy IQL emotion. 
Among the top-$25\%$ judge-filtered turns used for emotion-free SFT, only $25/526$ turns ($4.8\%$) match. 
Therefore, the emotion-free student cannot recover the intended emotional mode from the dialogue state alone.
This explains the empirical pattern in Panel B of Table~\ref{tab:mechanism_ablation}: emotion-free training can produce strong agreements when successful, but its reliability drops because the explicit emotion control variable is removed. 
Emotion-conditional EmoDistill remains the default deployment setting, while emotion-free EmoDistill is best interpreted as a diagnostic ablation for testing whether emotional strategy can be internalized without an explicit emotion channel.

\section{Bootstrap Confidence Intervals for Outcomes}
\label{app:bootstrap_ci}

This appendix reports $95\%$ bootstrap confidence intervals for
\emph{Outcomes} in the in-domain experiments and CRAD training-signal ablation, as shown in
Tables~\ref{tab:ci_outcome}and
\ref{tab:ci_outcomes_reward_crad}. Outcomes are computed as the average role-neutral signed
savings over successful negotiations. We provide intervals for the in-domain
experiments, the CRAD training-signal ablation, and the CRAD cross-counterparty
transfer setting.

\paragraph{Procedure.}
For each method--condition cell, we first collect the successful held-out
episodes,
\[
\mathcal{S} = \{i : \text{episode } i \text{ succeeds}\}.
\]
For each successful episode $i \in \mathcal{S}$, we compute
\[
o_i = \textsc{Sav}_i,
\]
where $\textsc{Sav}_i$ is the role-neutral signed savings defined in
Eq.~\eqref{eq:sav_signinv} and clipped to $[0,1]$. Let
$N_{\mathrm{succ}} = |\mathcal{S}|$ be the number of successful episodes for
that cell. We draw $B{=}10{,}000$ percentile-bootstrap resamples of size
$N_{\mathrm{succ}}$ with replacement from
$\{o_i : i \in \mathcal{S}\}$. For each resample, we compute the mean Outcome
score. The reported confidence interval is given by the $2.5\%$ and $97.5\%$
percentiles of the bootstrap distribution.
If a method has no successful episodes in a cell, the Outcomes confidence
interval is undefined and marked as N/A.

\paragraph{Interpreting the intervals.}
The confidence intervals quantify uncertainty in successful-deal quality,
conditional on the negotiation reaching agreement. They should therefore be
interpreted together with the corresponding success rates. A method may have a
high Outcome score with a wide interval if only a small or heterogeneous set of
successful episodes contributes to the estimate. Conversely, high success with
a narrow Outcome interval indicates more stable agreement quality across
held-out scenarios.
Because Outcomes are computed only over successful episodes, these intervals
measure agreement quality rather than agreement probability. They are therefore
intended as a complementary uncertainty analysis for successful negotiation
outcomes.

\begin{table*}[t]
\centering
\fontsize{9}{10}\selectfont
\setlength{\tabcolsep}{4pt}
\renewcommand{\arraystretch}{1.05}
\scalebox{0.92}{%
\begin{tabular}{p{1.55cm}lcc}
\toprule
\textbf{Dataset} & \textbf{Method} & \textbf{Outcomes} (\%) & \textbf{95\% CI} \\
\midrule

\multirow{7}{1.55cm}{CRAD}
& Vanilla (LLM) & $10.0$ & $[0.0,\,23.4]$ \\
& Vanilla (SLM) & $35.2$ & $[11.1,\,59.3]$ \\
& Random & $47.7$ & $[31.7,\,54.2]$ \\
\cmidrule(lr){2-4}
& IQL & $66.9$ & $[52.1,\,80.3]$ \\
& IQL+SFT & $69.8$ & $[56.5,\,83.1]$ \\
& IQL+JPO & $54.4$ & $[39.7,\,69.2]$ \\
& \textbf{IQL+SFT+JPO} & $\textbf{80.2}$ & $\textbf{[66.4,\,94.0]}$ \\

\midrule
\multirow{7}{1.55cm}{Disaster}
& Vanilla (LLM) & $15.0$ & $[0.0,\,30.6]$ \\
& Vanilla (SLM) & $50.5$ & $[29.8,\,71.3]$ \\
& Random & $10.0$ & $[0.0,\,23.1]$ \\
\cmidrule(lr){2-4}
& IQL & $5.0$ & $[0.0,\,14.6]$ \\
& IQL+SFT & $15.0$ & $[0.0,\,30.6]$ \\
& IQL+JPO & $\textbf{44.4}$ & $\textbf{[21.5,\,67.4]}$ \\
& IQL+SFT+JPO & $30.0$ & $[9.9,\,50.1]$ \\

\midrule
\multirow{7}{1.55cm}{Hospital}
& Vanilla (LLM) & $35.0$ & $[14.1,\,55.9]$ \\
& Vanilla (SLM) & $44.8$ & $[23.9,\,65.7]$ \\
& Random & $30.0$ & $[9.9,\,50.1]$ \\
\cmidrule(lr){2-4}
& IQL & $0.0$ & $[0.0,\,0.0]$ \\
& IQL+SFT & $21.1$ & $[2.7,\,39.4]$ \\
& IQL+JPO & $35.0$ & $[14.1,\,55.9]$ \\
& \textbf{IQL+SFT+JPO} & $\textbf{45.0}$ & $\textbf{[23.2,\,66.8]}$ \\

\midrule
\multirow{7}{1.55cm}{Student}
& Vanilla (LLM) & $45.9$ & $[36.5,\,55.3]$ \\
& Vanilla (SLM) & $15.0$ & $[1.5,\,28.5]$ \\
& Random & $43.8$ & $[30.4,\,57.2]$ \\
\cmidrule(lr){2-4}
& IQL & $47.9$ & $[34.5,\,61.3]$ \\
& IQL+SFT & $51.7$ & $[42.6,\,60.8]$ \\
& IQL+JPO & $23.5$ & $[8.6,\,38.4]$ \\
& \textbf{IQL+SFT+JPO} & $\textbf{52.6}$ & $\textbf{[40.9,\,64.3]}$ \\

\bottomrule
\end{tabular}%
}
\caption{Bootstrap $95\%$ confidence intervals for successful-case \textbf{Outcomes} in the in-domain experiments. Outcomes are computed over successful held-out episodes only. For each method--dataset cell, we resample successful-episode $\textsc{Sav}_i$ values with replacement using $10{,}000$ percentile-bootstrap resamples and report the $2.5\%$ and $97.5\%$ percentiles. EmoDistill denotes the full IQL+SFT+JPO pipeline.}
\label{tab:ci_outcome}
\end{table*}

\begin{table*}[t]
\centering
\fontsize{9}{10}\selectfont
\setlength{\tabcolsep}{6pt}
\renewcommand{\arraystretch}{1.05}
\scalebox{0.95}{%
\begin{tabular}{lcc}
\toprule
\textbf{Signal} & \textbf{Outcomes} (\%) & \textbf{95\% CI} \\
\midrule
SFT: hybrid quality filter & $69.8$ & $[55.6,\,84.0]$ \\
SFT: episode-judge filter & $55.1$ & $[38.5,\,71.7]$ \\
SFT: turn-judge filter & $64.8$ & $[49.4,\,80.2]$ \\
\midrule
JPO: outcome-shaped advantage & $64.1$ & $[49.7,\,78.5]$ \\
JPO: episode-judge advantage & $77.4$ & $[64.5,\,90.3]$ \\
\textbf{JPO: turn-judge advantage} & $\textbf{80.2}$ & $\textbf{[65.1,\,95.3]}$ \\
\bottomrule
\end{tabular}%
}
\caption{Bootstrap $95\%$ confidence intervals for the \textbf{Outcomes} column of Panel A of Table~\ref{tab:mechanism_ablation} on the CRAD training-signal ablation. Outcomes are computed over successful negotiations only. For each method--signal cell, we resample successful-episode $\textsc{Sav}_i$ values with replacement using $10{,}000$ percentile-bootstrap resamples and report the $2.5\%$ and $97.5\%$ percentiles. These intervals quantify uncertainty in successful-deal quality for different training signals.}
\label{tab:ci_outcomes_reward_crad}
\end{table*}

\section{Emotion-Free Distillation Across Domains}
\label{app:emotion_free}

Table~\ref{tab:prompt_free_ablation_full} shows that emotion-free distillation is not simply a failed version of EmoDistill, but its behavior is strongly domain- and stage-dependent.
Removing the explicit emotion channel can still allow the LoRA adapter to absorb useful bargaining behavior from high-quality offline trajectories, yet the best emotion-free training stage differs across domains.
On CRAD, emotion-free SFT substantially improves over Vanilla (SLM), increasing Utility from $8.8$ to $55.0$ with a success rate of $90.0\%$.
Adding JPO increases successful-case Outcomes to $76.7$, the highest among the CRAD emotion-free variants, but success drops to $50.0\%$, reducing Utility to $38.4$.
The pattern is different on Disaster.
Here, SFT alone does not help: Utility falls from the Vanilla SLM baseline of $37.9$ to $10.0$.
However, SFT+JPO recovers strongly, reaching $100.0\%$ success and the highest Utility, $64.2$.

Hospital is the weakest setting for emotion-free distillation.
Vanilla (SLM) remains the best method by Utility, with $40.3$, while SFT and SFT+JPO underperform substantially.
Direct JPO without SFT reaches a similar Utility of $40.0$, but with lower success and much longer negotiations.
On Student, emotion-free SFT again provides a strong generic bargaining policy, improving Utility from $15.0$ to $47.6$.
SFT+JPO obtains the highest successful-case Outcomes, $51.0$, but its lower success rate reduces Utility to $45.9$, slightly below SFT.
Overall, emotion-free training can internalize useful negotiation behavior, and the best emotion-free variant improves over Vanilla (SLM) on three of four domains.
However, the gains are not uniform: SFT is strongest on CRAD and Student, SFT+JPO is strongest on Disaster, and no learned emotion-free variant clearly improves over Vanilla on Hospital.
This reinforces the role of the explicit emotion channel in EmoDistill.

\begin{table*}[t]
\centering
\setlength{\cmidrulewidth}{0.2pt}
\setlength{\aboverulesep}{1.5pt}
\setlength{\belowrulesep}{1.5pt}
\renewcommand{\arraystretch}{1.05}
\caption{Prompt-free ablation across four datasets (expansion of Panel B of Table~\ref{tab:mechanism_ablation}). \textbf{Setup}: focal/creditor = Qwen2.5-7B + LoRA (SLM), counterparty/debtor = Qwen3.5-Plus (LLM). All methods remove the emotion block from both training and inference. Disaster and Hospital savings use the sign-invariant metric (Appendix~\ref{app:normalized_savings}). \textbf{Success}~$\uparrow$, \textbf{Outcomes} = mean$\pm$std over successful episodes, \textbf{Utility} = mean$\pm$std over all 20 episodes (failures=0), \textbf{Rounds}~$\downarrow$. Best \textsc{Utility} per dataset is in \textbf{bold}; best successful-case \textsc{Outcomes} are also bolded when discussed. }
\label{tab:prompt_free_ablation_full}
\scalebox{0.78}{%
\begin{tabular}{p{1.5cm}lcccc}
\toprule
\textbf{Dataset} & \textbf{Method} & \textbf{Success (\%)} $\uparrow$ & \textbf{Outcomes} (\%) $\uparrow$ & \textbf{Utility} (\%) $\uparrow$ & \textbf{Rounds} $\downarrow$ \\
\midrule
\multirow{4}{1.5cm}{CRAD (Debt)}
& Vanilla (SLM, baseline)          & 25.0  & $35.3{\scriptstyle\pm26.1}$ & $8.8{\scriptstyle\pm20.1}$ & $11.2{\scriptstyle\pm5.9}$ \\
& SFT (no emotion)                 & 90.0  & $61.1{\scriptstyle\pm28.8}$ & $\mathbf{55.0{\scriptstyle\pm32.5}}$ & $13.7{\scriptstyle\pm10.2}$ \\
& JPO (no emotion, no SFT)         & 40.0  & $28.7{\scriptstyle\pm30.6}$ & $11.5{\scriptstyle\pm23.4}$ & $24.2{\scriptstyle\pm9.6}$ \\
& \textbf{SFT+JPO} (ours, no emo)  & 50.0  & $\mathbf{76.7{\scriptstyle\pm32.2}}$ & $38.4{\scriptstyle\pm46.2}$ & $23.6{\scriptstyle\pm9.1}$ \\
\midrule
\multirow{4}{1.5cm}{Disaster (Rescue)}
& Vanilla (SLM, baseline)          & 75.0  & $50.5{\scriptstyle\pm41.0}$ & $37.9{\scriptstyle\pm41.7}$ & $9.9{\scriptstyle\pm7.2}$ \\
& SFT (no emotion)                 & 95.0  & $10.5{\scriptstyle\pm31.5}$ & $10.0{\scriptstyle\pm30.8}$ & $7.5{\scriptstyle\pm7.7}$ \\
& JPO (no emotion, no SFT)         & 70.0  & $21.4{\scriptstyle\pm42.6}$ & $15.0{\scriptstyle\pm36.6}$ & $15.9{\scriptstyle\pm11.6}$ \\
& \textbf{SFT+JPO} (ours, no emo)  & 100.0 & $\mathbf{64.2{\scriptstyle\pm47.2}}$ & $\mathbf{64.2{\scriptstyle\pm47.2}}$ & $4.2{\scriptstyle\pm2.0}$ \\
\midrule
\multirow{4}{1.5cm}{Hospital (Medical)}
& Vanilla (SLM, baseline)          & 90.0  & $\mathbf{44.7{\scriptstyle\pm45.3}}$ & $\mathbf{40.3{\scriptstyle\pm45.0}}$ & $4.3{\scriptstyle\pm2.8}$ \\
& SFT (no emotion)                 & 100.0 & $10.0{\scriptstyle\pm30.8}$ & $10.0{\scriptstyle\pm30.8}$ & $3.7{\scriptstyle\pm1.6}$ \\
& JPO (no emotion, no SFT)         & 85.0  & $47.1{\scriptstyle\pm51.4}$ & $40.0{\scriptstyle\pm50.3}$ & $9.9{\scriptstyle\pm11.1}$ \\
& \textbf{SFT+JPO} (ours, no emo)  & 100.0 & $23.7{\scriptstyle\pm35.5}$ & $23.7{\scriptstyle\pm35.5}$ & $3.4{\scriptstyle\pm1.9}$ \\
\midrule
\multirow{4}{1.5cm}{Student (Sleep)}
& Vanilla (SLM, baseline)          & 100.0 & $15.0{\scriptstyle\pm30.7}$ & $15.0{\scriptstyle\pm30.7}$ & $3.4{\scriptstyle\pm1.5}$ \\
& SFT (no emotion)                 & 95.0  & $50.1{\scriptstyle\pm22.5}$ & $47.6{\scriptstyle\pm23.4}$ & $3.2{\scriptstyle\pm2.3}$ \\
& JPO (no emotion, no SFT)         & 90.0  & $2.6{\scriptstyle\pm7.5}$ & $2.2{\scriptstyle\pm7.0}$ & $9.3{\scriptstyle\pm10.1}$ \\
& \textbf{SFT+JPO} (ours, no emo)  & 90.0  & $\mathbf{51.0{\scriptstyle\pm22.8}}$ & $\mathbf{45.9{\scriptstyle\pm26.2}}$ & $2.2{\scriptstyle\pm0.7}$ \\
\bottomrule
\end{tabular}%
}
\end{table*}

\begin{table}[t]
\centering
\setlength{\tabcolsep}{1.2pt}
\setlength{\cmidrulewidth}{0.2pt}
\setlength{\aboverulesep}{1.5pt}
\setlength{\belowrulesep}{1.5pt}
\renewcommand{\arraystretch}{1.08}
\caption{Trained-vs-trained tournament on \textbf{CRAD}. V denotes Vanilla, I denotes IQL, E denotes IQL+SFT+JPO, and E$_0$ denotes the condition-free variant. Each cell reports Success rate (\%) / Utility, with failures counted as 0. Best Utility per counterparty is in \textbf{bold}.}
\label{tab:tournament}

\begingroup
\fontsize{7.5}{8.5}\selectfont
\begin{tabular*}{\columnwidth}{@{\extracolsep{\fill}}l|cccc@{}}
\toprule
\textbf{F$\backslash$C} & \textbf{V} & \textbf{I} & \textbf{E} & \textbf{E$_0$} \\
\midrule
\textbf{V}
& 5.0/$3.2_{\pm13.7}$
& 5.0/$3.2_{\pm13.7}$
& 5.0/$5.0_{\pm21.8}$
& 15.0/$11.1_{\pm27.3}$ \\

\textbf{I}
& 5.0/$3.2_{\pm13.7}$
& 5.0/$3.2_{\pm13.7}$
& 5.0/$5.0_{\pm21.8}$
& 15.0/$11.1_{\pm27.3}$ \\

\textbf{E}
& 10.0/$\bm{9.7_{\pm29.0}}$
& 10.0/$\bm{9.7_{\pm29.0}}$
& 15.0/$\bm{14.4_{\pm34.2}}$
& 5.0/$4.7_{\pm20.4}$ \\

\textbf{E$_0$}
& 5.0/$5.0_{\pm21.8}$
& 5.0/$5.0_{\pm21.8}$
& 15.0/$\bm{14.4_{\pm34.0}}$
& 15.0/$\bm{15.0_{\pm35.7}}$ \\
\bottomrule
\end{tabular*}
\endgroup
\end{table}

\begin{table*}[t]
\centering
\fontsize{9}{10}\selectfont
\setlength{\tabcolsep}{4pt}
\renewcommand{\arraystretch}{1.05}
\caption{A-LoL vs. JPO refinement across four datasets. Both methods use the same IQL emotion selector and LoRA-SFT initialization; A-LoL applies positive-advantage refinement, while JPO uses judge-guided clipped refinement. Utility counts failures as 0. Best value per dataset is in \textbf{bold}, except Success.}
\label{tab:alol_jpo_compare}
\begin{tabular*}{\textwidth}{@{\extracolsep{\fill}}p{1.55cm}lcccc@{}}
\toprule
\textbf{Dataset} & \textbf{Method} 
& \textbf{Success (\%)} $\uparrow$ 
& \textbf{Outcomes} (\%) $\uparrow$ 
& \textbf{Utility} (\%) $\uparrow$ 
& \textbf{Rounds} $\downarrow$ \\
\midrule

\multirow{2}{1.55cm}{CRAD}
& A-LoL & 100.0 & $77.5{\scriptstyle\pm28.3}$ & $\mathbf{77.5{\scriptstyle\pm28.3}}$ & $\mathbf{8.7{\scriptstyle\pm4.6}}$ \\
& JPO   & 90.0  & $\mathbf{80.2{\scriptstyle\pm30.3}}$ & $72.2{\scriptstyle\pm37.5}$ & $15.0{\scriptstyle\pm9.8}$ \\

\midrule
\multirow{2}{1.55cm}{Disaster}
& A-LoL & 100.0 & $10.0{\scriptstyle\pm30.8}$ & $10.0{\scriptstyle\pm30.8}$ & $\mathbf{4.4{\scriptstyle\pm2.4}}$ \\
& JPO   & 100.0 & $\mathbf{30.0{\scriptstyle\pm45.8}}$ & $\mathbf{30.0{\scriptstyle\pm45.8}}$ & $6.5{\scriptstyle\pm4.0}$ \\

\midrule
\multirow{2}{1.55cm}{Hospital}
& A-LoL & 95.0  & $36.8{\scriptstyle\pm49.6}$ & $35.0{\scriptstyle\pm48.9}$ & $\mathbf{5.0{\scriptstyle\pm2.9}}$ \\
& JPO   & 100.0 & $\mathbf{45.0{\scriptstyle\pm49.7}}$ & $\mathbf{45.0{\scriptstyle\pm49.7}}$ & $5.5{\scriptstyle\pm3.4}$ \\

\midrule
\multirow{2}{1.55cm}{Student}
& A-LoL & 100.0 & $47.2{\scriptstyle\pm25.7}$ & $47.2{\scriptstyle\pm25.7}$ & $3.5{\scriptstyle\pm2.6}$ \\
& JPO   & 100.0 & $\mathbf{52.6{\scriptstyle\pm26.6}}$ & $\mathbf{52.6{\scriptstyle\pm26.6}}$ & $\mathbf{3.1{\scriptstyle\pm2.9}}$ \\

\bottomrule
\end{tabular*}
\end{table*}

\section{Positive-Advantage vs. Turn-Level Judge Refinement}
\label{app:alol_baseline}

A-LoL~\citep{baheti2024leftover} is a sequence-level offline RL baseline for language models. 
It first obtains an SFT reference policy, estimates an advantage for each prompt--response pair, and then applies advantage-weighted negative log-likelihood on positive-advantage examples. 
We include A-LoL as a refinement baseline after the same IQL selector and LoRA-SFT initialization used by EmoDistill. 
This comparison isolates the effect of the JPO update: A-LoL conservatively amplifies high-advantage SFT behavior, while JPO can use both positive- and negative-advantage utterances through a clipped objective and KL anchor. 
Moreover, JPO exposes a controllable risk parameter $\kappa$ that scales the pressure from negative-advantage samples. 
As shown in Table~\ref{tab:kappa_risk}, an intermediate setting, $\kappa=0.5$, achieves higher Utility than both canonical JPO and A-LoL on CRAD, suggesting that controlled negative-sample pressure can improve the success--value tradeoff.
The direct A-LoL/JPO comparison is shown in Table~\ref{tab:alol_jpo_compare}.

\begin{table}[t]
\centering
\setlength{\tabcolsep}{3.5pt}
\renewcommand{\arraystretch}{0.95}

\begingroup
\fontsize{7.8}{8.2}\selectfont

\begin{tabular*}{\columnwidth}{@{\extracolsep{\fill}}lccc@{}}
\toprule
\multicolumn{4}{l}{\textit{A. Heterogeneous counterparty robustness}} \\
\midrule
\textbf{Full vs.}
& \textbf{$\Delta U$}
& \textbf{95\% CI}
& \textbf{$p$} \\
\midrule

DeepSeek--Vanilla
& +41.1 & $[+22.1,+59.4]$ & .0014 \\

DeepSeek--IQL
& +49.5 & $[+36.1,+62.4]$ & $<10^{-4}$ \\

DeepSeek--IQL+SFT
& +14.7 & $[+1.0,+28.9]$ & .0539 \\

GPT-4o--Vanilla
& +33.3 & $[+13.3,+53.1]$ & .0124 \\

GPT-4o--IQL
& +24.8 & $[+13.6,+37.5]$ & .0001 \\

GPT-4o--IQL+SFT
& +8.9 & $[-1.7,+19.1]$ & .0728 \\

\bottomrule
\end{tabular*}

\vspace{3pt}

\begin{tabular*}{\columnwidth}{@{\extracolsep{\fill}}lc@{}}
\toprule
\multicolumn{2}{l}{\textit{B. Reward-labeler robustness}} \\
\midrule
\textbf{Evidence}
& \textbf{Result} \\
\midrule

DeepSeek-labeled \textsc{EmoDistill}
& 100.0 Suc. / 63.5 Util. \\

Qwen- vs.\ DeepSeek-labeled
& $+5.4$ Util., $p=.4653$ \\

DeepSeek-labeled vs.\ Vanilla
& $+35.7$ Util., $p=.0003$ \\

DeepSeek-labeled vs.\ Random
& $+30.7$ Util., $p=.0023$ \\

DeepSeek-labeled vs.\ IQL
& $+44.1$ Util., $p=.0007$ \\

\bottomrule
\end{tabular*}

\caption{
Robustness checks on \textbf{CRAD}.
\textbf{A:} Paired Utility comparisons of \textsc{EmoDistill} against
baselines under unseen DeepSeek-V3 and GPT-4o counterparties;
95\% confidence intervals are computed for the paired Utility differences.
\textbf{B:} Reward-labeler robustness after re-labeling the same 4,000 JPO
training examples with DeepSeek-V3.
}
\label{tab:judge_counterparty_robustness}

\endgroup
\end{table}

\section{Judge and Counterparty Robustness}
\label{app:judge_counterparty_robustness}

The main experiments use Qwen-based offline trajectories and reward annotations, which raises the possibility that the observed gains are specific to a particular judge or counterparty.
We therefore perform two additional robustness checks on CRAD.

First, we evaluate the same trained policies against two unseen large-model counterparties, DeepSeek-V3 and GPT-4o, using the same 20 held-out scenarios and report paired Utility differences relative to the full \textsc{EmoDistill} model.
Second, we re-label the same 4,000 JPO training examples with DeepSeek-V3 and retrain the refinement stage, directly testing sensitivity to the reward-labeler model.

As shown in Table~\ref{tab:judge_counterparty_robustness}, the full model retains substantial paired Utility gains over Vanilla and IQL under both unseen counterparties.
The incremental advantage over IQL+SFT is smaller and not statistically significant under either DeepSeek-V3 or GPT-4o, so we do not claim that JPO uniformly improves every counterparty setting.

Changing the reward labeler also substantially alters the preference signal: DeepSeek agrees with the original Qwen ordering on only $59.3\%$ of the pairs and reverses $36.9\%$.
Nevertheless, the DeepSeek-labeled policy achieves $100\%$ Success and $63.5$ Utility against DeepSeek-V3, and its $5.4$-point Utility difference from the Qwen-labeled policy is not significant ($p=.4653$).
Together, these results suggest that the gains are not tied to a single judge or large-model counterparty, although the smaller-counterparty experiments reveal a separate scale-related boundary condition.

\subsection{Smaller-Counterparty Boundary}
\label{app:small_counterparty}

The large-model counterparty results do not imply universal transfer.
We therefore evaluate the same CRAD-trained policies against
Qwen2.5-3B-Instruct, a substantially smaller counterparty.

\begin{table}[t]
\centering
\setlength{\tabcolsep}{3pt}
\renewcommand{\arraystretch}{0.95}

\begingroup
\fontsize{7.8}{8.2}\selectfont
\begin{tabular*}{\columnwidth}
{@{\extracolsep{\fill}}lcccc@{}}
\toprule
\textbf{Method}
& \textbf{Suc.}
& \textbf{Out.}
& \textbf{Util.}
& \textbf{Rd.} \\
\midrule

Vanilla 7B
& 30.0
& $60.1{\scriptstyle\pm40.5}$
& $18.0{\scriptstyle\pm35.4}$
& $4.5{\scriptstyle\pm2.1}$ \\

IQL 7B
& 85.0
& $\mathbf{73.0{\scriptstyle\pm34.7}}$
& $\mathbf{62.1{\scriptstyle\pm41.3}}$
& $9.8{\scriptstyle\pm9.5}$ \\

IQL+SFT 7B
& 30.0
& $57.4{\scriptstyle\pm30.4}$
& $17.2{\scriptstyle\pm31.1}$
& $7.0{\scriptstyle\pm10.4}$ \\

\textbf{\textsc{EmoDistill}}
& 25.0
& $53.1{\scriptstyle\pm35.4}$
& $13.3{\scriptstyle\pm29.0}$
& $8.7{\scriptstyle\pm12.5}$ \\

\bottomrule
\end{tabular*}

\caption{
Smaller-counterparty boundary on \textbf{CRAD} with
Qwen2.5-3B-Instruct.
The full \textsc{EmoDistill} policy transfers substantially less
reliably than under DeepSeek-V3 and GPT-4o, indicating that
cross-counterparty transfer is scale dependent.
}
\label{tab:small_counterparty}
\endgroup
\end{table}

Unlike the transfer observed with DeepSeek-V3 and GPT-4o,
the full model reaches only $25.0\%$ Success and $13.3$ Utility
against Qwen2.5-3B-Instruct.
IQL alone performs substantially better in this setting.
We therefore treat the 3B result as a boundary condition:
the expression patterns distilled from large-model negotiations
do not transfer uniformly to substantially smaller counterparties.

\section{Stronger Offline-Alignment Baselines}
\label{app:stronger_offline_baselines}

We additionally compare \textsc{EmoDistill} with DPO, ORPO, and KTO
under a matched offline-alignment setting.
All methods use the same Qwen2.5-7B-Instruct backbone, offline CRAD
training data, and LoRA parameter budget.
We evaluate two values of $\beta$ for each baseline on
$100$ additional, newly generated held-out CRAD scenarios,
disjoint from training, tuning, and model selection.
Evaluation uses DeepSeek-V3 as the counterparty.

\begin{table}[t]
\centering
\setlength{\tabcolsep}{3.5pt}
\renewcommand{\arraystretch}{0.95}

\begingroup
\fontsize{7.8}{8.2}\selectfont
\begin{tabular*}{\columnwidth}
{@{\extracolsep{\fill}}lcccc@{}}
\toprule
\textbf{Method}
& \textbf{Util.}
& \textbf{$\Delta U$}
& \textbf{95\% CI}
& \textbf{$p$} \\
\midrule

DPO ($\beta=.01$)
& 47.0 & +18.2 & $[+10.8,+25.4]$ & $<10^{-4}$ \\

DPO ($\beta=.001$)
& 53.1 & +12.1 & $[+4.5,+19.8]$ & .0031 \\

ORPO ($\beta=.01$)
& 58.3 & +6.9 & $[-0.5,+14.2]$ & .0920 \\

ORPO ($\beta=.001$)
& 53.4 & +11.7 & $[+5.0,+18.5]$ & .0005 \\

KTO ($\beta=.01$)
& 39.8 & +25.3 & $[+17.4,+32.8]$ & $<10^{-4}$ \\

KTO ($\beta=.001$)
& 57.6 & +7.5 & $[+0.1,+15.1]$ & .0712 \\

\midrule
\textbf{\textsc{EmoDistill}}
& \textbf{65.1} & -- & -- & -- \\

\bottomrule
\end{tabular*}

\caption{
Stronger offline-alignment baselines on $100$ additional held-out
\textbf{CRAD} scenarios against DeepSeek-V3.
$\Delta U$ denotes the paired Utility improvement of
\textsc{EmoDistill} over each baseline.
The final column reports paired Wilcoxon tests.
}
\label{tab:strong_offline_appendix}
\endgroup
\end{table}

\textsc{EmoDistill} achieves $99.0\%$ Success and $65.1$ Utility.
Its paired Utility difference is positive against every evaluated
configuration.
The gains over both DPO settings, ORPO ($\beta=.001$), and
KTO ($\beta=.01$) are statistically significant, whereas the gains
over ORPO ($\beta=.01$) and KTO ($\beta=.001$) are positive but not
conventionally significant.
We therefore do not claim uniform statistical superiority.

\section{Training Stability Analysis}
\label{app:training_stability}

This appendix reports optimization stability diagnostics for the three learned components in EmoDistill: the IQL emotion selector, the LoRA-SFT initializer, and the JPO refiner. 
These diagnostics do not prove global convergence; they check whether each stage remains numerically stable under fixed offline training. 
For each logged quantity $\{\ell_t\}_{t=1}^{T}$, we summarize the final $25\%$ of training by its median $\widetilde{\ell}$ and robust dispersion $\mathrm{MAD}=\mathrm{median}|\ell_t-\widetilde{\ell}|$. 
A stable run should show bounded late-stage variation and no uncontrolled drift.

\paragraph{IQL selector.}
For the offline emotion selector, we track the V-network expectile loss $\mathcal{L}_V$, Q-network TD loss $\mathcal{L}_Q$, and AWR policy loss $\mathcal{L}_{\pi}$ (Eqs.~\eqref{eq:iql_vq}--\eqref{eq:iql_pi}). 
On CRAD, $\mathcal{L}_Q$ decreases from roughly $0.9$ to $0.32$, $\mathcal{L}_V$ plateaus near $0.12$, and the AWR policy loss stabilizes around $2.8$. 
In the final quarter, the medians are $\widetilde{\mathcal{L}_V}=0.115$, $\widetilde{\mathcal{L}_Q}=0.323$, and $\widetilde{\mathcal{L}_{\pi}}=2.835$, with bounded MADs. 
This indicates that the selector's value estimates are stable before policy extraction.

\paragraph{LoRA-SFT initialization.}
Stage~1 fine-tunes a rank-$16$ LoRA adapter on the top-$25\%$ hybrid-filtered subset of the CRAD sweep. 
The token-level cross-entropy decreases from $2.15$ to $0.69$ over two epochs and reaches a stable plateau in the second epoch. 
The final-quarter median is $\widetilde{\mathcal{L}_{\mathrm{SFT}}}=0.711$ with MAD $0.003$, giving JPO a stable reference policy $\pi_{\mathrm{ref}}$.

\paragraph{JPO refinement.}
For JPO, the key diagnostics are the in-training reward objective $-\mathcal{L}_{\mathrm{PG}}=\mathbb{E}[\rho_t A_t]$, KL divergence to the frozen SFT reference, and the importance ratio $\rho_t$. 
We trained $5$-epoch emotion-free JPO runs on CRAD, Disaster, and Student to expose cross-epoch behavior. 
As shown in Table~\ref{tab:training_stability}, the reward signal improves across datasets, while the control metrics remain stable: $\widetilde{\rho_t}\in[0.89,0.94]$ with MAD $\leq0.006$, and $\widetilde{\mathrm{KL}}\in[0.14,0.23]$ with MAD $\leq0.010$. 
Across $493$ logged JPO points, we observe zero importance-ratio clip violations, indicating that JPO stays within the intended trust region.

\begin{table*}[t]
\centering
\fontsize{9}{10}\selectfont
\setlength{\tabcolsep}{4pt}
\renewcommand{\arraystretch}{1.1}
\caption{\textbf{Training stability summary.} For each row, we report logged points / total steps, first-quartile vs. last-quartile signal, last-quartile median (MAD) of the control metric, and violation counts. For JPO, reward is $-\mathcal{L}_{\mathrm{PG}}$; KL spikes count $\mathrm{KL}>0.5$ events; $\rho_t$ clips count points outside $[0.8,1.2]$.}
\label{tab:training_stability}
\scalebox{0.93}{%
\begin{tabular}{llccccc}
\toprule
\textbf{Stage} & \textbf{Dataset} & \textbf{Pts / Steps} & \textbf{Signal: q1 $\to$ q4} & \textbf{Last-q median (MAD)} & \textbf{KL spikes} & \textbf{$\rho_t$ clips} \\
\midrule
IQL & CRAD & 1001 / 50{,}000 & $\mathcal{L}_Q$: $0.48 \to 0.32$ & $\widetilde{\mathcal{L}_Q}{=}0.323$ (0.043) & n/a & n/a \\
LoRA-SFT & CRAD & 25 / 625 & $\mathcal{L}_{\mathrm{SFT}}$: $1.09 \to 0.71$ & $\widetilde{\mathcal{L}_{\mathrm{SFT}}}{=}0.711$ (0.003) & n/a & n/a \\
\midrule
JPO & CRAD & 263 / 6{,}575 & $-\mathrm{pg}$: $+0.18 \to +0.21$ & $\widetilde{\mathrm{KL}}{=}0.23$ (0.010), $\widetilde{\rho_t}{=}0.94$ (0.005) & 4 / 263 (1.5\%) & \textbf{0} \\
JPO & Disaster & 146 / 3{,}650 & $-\mathrm{pg}$: $-0.01 \to +0.03$ & $\widetilde{\mathrm{KL}}{=}0.19$ (0.004), $\widetilde{\rho_t}{=}0.89$ (0.006) & 4 / 146 (2.7\%) & \textbf{0} \\
JPO & Student & 84 / 2{,}100 & $-\mathrm{pg}$: $+0.05 \to +0.06$ & $\widetilde{\mathrm{KL}}{=}0.14$ (0.003), $\widetilde{\rho_t}{=}0.90$ (0.005) & 0 / 84 (0.0\%) & \textbf{0} \\
\bottomrule
\end{tabular}%
}
\end{table*}

\paragraph{Interpretation.}
These diagnostics show that EmoDistill is numerically stable under offline training. 
The IQL selector stabilizes before policy extraction, the SFT adapter provides a reliable reference distribution, and JPO improves judge-aligned behavior while keeping KL and importance ratios controlled. 
This is important because JPO is trained on a fixed offline sweep; without bounded KL and controlled $\rho_t$, the policy could exploit fixed judge labels out of distribution.

\section{Implementation}
\label{app:hardware}

All \textsc{EmoDistill} experiments were run on a single workstation with
$4\times$ NVIDIA RTX 4090 GPUs (24\,GB each), Ubuntu 22.04, CUDA 12.4, and
PyTorch 2.4. The Qwen2.5-7B-Instruct student is trained with rank-16 LoRA
adapters in bf16 mixed precision; each training process fits within one 24\,GB
GPU. We parallelize experiments by dataset, assigning CRAD, Disaster, Hospital,
and Student to separate GPUs without inter-GPU gradient synchronization. The
full set of reported experiments, including main results, ablations, transfer
tests, and the $\kappa$ sweep, required approximately $42$ GPU-hours on RTX
4090 GPUs.

\section{Cost Analysis}
\label{app:cost_analysis}

We estimate the API cost of reproducing the four-dataset
\textsc{EmoDistill} pipeline using Qwen3.5-Plus at the public list
price used during our experiments.
The main cost comes from three stages: generating the offline
LLM-vs-LLM negotiation sweep, annotating focal-agent turns with the
LLM judge, and running held-out evaluation with the default
API-served counterparty.
Across CRAD, Disaster, Hospital, and Student, the default
Qwen3.5-Plus pipeline costs approximately \$44.6 in total:
about \$31.0 for the two-sided sweep, \$10.1 for per-turn judge
annotation, and \$3.5 for held-out evaluation.
This estimate excludes the additional DeepSeek-V3/GPT-4o
cross-counterparty evaluations and the DeepSeek-V3 reward-labeler
robustness experiment, whose costs depend on the corresponding API
pricing.
Local GPU compute is reported separately in
Appendix~\ref{app:hardware}.

\section{Use of Large Language Models}
\label{app:llm_use}

LLMs are used in this paper as experimental components and as limited
writing assistants.
Qwen3.5-Plus is used to generate the default offline LLM-vs-LLM
negotiation sweep, provide the default per-turn reward annotations,
and serve as the default in-domain counterparty.
The distilled student negotiator is Qwen2.5-7B-Instruct with LoRA
adapters.
Cross-counterparty experiments additionally evaluate DeepSeek-V3,
GPT-4o, and Qwen2.5-3B-Instruct, while the reward-labeler robustness
experiment re-labels the JPO data with DeepSeek-V3.
LLM-judge scores are used for training-time annotation, demonstration
filtering, and JPO refinement, but final Success, Outcomes, and Utility
are computed from terminal agreements and predefined quantitative
objectives.
For manuscript preparation, the authors used LLMs only for grammar and
figure polishing, sentence-level rephrasing, LaTeX cleanup, and
table/caption formatting.
All technical claims, method design choices, derivations, experimental
analyses, and interpretations were authored and verified by the human
authors.

\section{Prompts}
\label{app:prompts}

This appendix documents the four prompt families that define the input/output interface of EmoDistill: the focal-agent system prompt, the counterparty system prompt, the per-turn LLM-judge prompt, and the per-emotion conditioning block inserted into the focal-agent prompt. Together with the four scenario CSVs (Appendix~\ref{app:dataset_details}) and the offline-sweep specification (Appendix~\ref{app:sweep_construction}), these prompts are sufficient to reproduce the methods and results in this paper.
Curly-brace placeholders such as \texttt{\{target\_days\}} and \texttt{\{outstanding\_balance\}} are filled from the structured scenario record. The focal agent's emotion block is the only \emph{inference-time control variable}; all other prompt components are fixed within each dataset. During training, the emotion block is sampled uniformly from the $|\mathcal{E}|=28$ action vocabulary (Appendix~\ref{app:sweep_construction}); during deployment, it is chosen by the IQL selector. The Emotion-Free variant (\S\ref{subsec:method_variants}) removes this block entirely.

\subsection{Example 1: Focal-agent system prompt (CRAD creditor)}
\label{app:prompts_creditor}

This is the system message the focal agent (creditor on CRAD) receives at every turn of a negotiation. It establishes four things in a fixed order: (i)~the strategic rules of the negotiation game (no copying the counterparty's exact number, gradual movement, minimization objective); (ii)~role-clarity instructions (no role labels in output, $1$--$2$ sentence response cap); (iii)~scenario-specific context loaded from the structured scenario record (outstanding balance, focal target, recovery stage, business context); and (iv)~the dialogue history rendered as a timeline string. The very last block before the response cue is the \texttt{\{emotion\_text\}} slot --- the only inference-time control variable for the focal agent. Both during the offline sweep (where the emotion is sampled uniformly from $\mathcal{E}$) and at deployment (where it is chosen by the IQL selector), this is the slot whose contents are swapped to change the agent's behavior; all other prompt components stay byte-identical within a scenario. The Emotion-Free ablation removes the entire ``EMOTIONAL APPROACH'' block at evaluation time via the environment variable \texttt{EVOEMO\_NO\_EMOTION\_PROMPT=1}.

\begin{figure*}[!t]
\centering
\begin{tcolorbox}[colback=gray!3, colframe=gray!40, title=\textbf{CRAD creditor prompt --- focal-agent system message}, fonttitle=\small, fontupper=\footnotesize]
\textit{Used by:} all CRAD focal-agent (creditor) calls during sweep and evaluation. \quad \textit{Placeholders:} \texttt{\{outstanding\_balance\}}, \texttt{\{target\_days\}}, \texttt{\{recovery\_stage\}}, \texttt{\{business\_sector\}}, \texttt{\{reason\_for\_overdue\}}, \texttt{\{timeline\_text\}}, \texttt{\{emotion\_text\}}.

\smallskip\hrule\smallskip
\begin{Verbatim}[fontsize=\scriptsize]
You are a PROFESSIONAL Creditor debt collection agent negotiating
payment timeline with the Debtor.

### CRITICAL NEGOTIATION RULES:
- NEVER copy the debtor's exact number - this shows weakness
- Move GRADUALLY toward their position (not all at once)
- Show you are negotiating, not just accepting
- Your goal: Minimize payment days while reaching agreement

### ROLE CLARITY
- You are ONLY the Creditor - speak only as yourself
- Do NOT include "**Creditor:**" or "**Debtor:**" labels
- Give only YOUR response as the creditor (1-2 sentences max)

### DEBT COLLECTION CONTEXT
- Outstanding Balance: ${outstanding_balance}
- Your Target Timeline: {target_days} days for full payment
- Recovery Stage: {recovery_stage}
- Debtor Business: {business_sector}
- Reason for Overdue: {reason_for_overdue}

### CURRENT SITUATION
{timeline_text}

### EMOTIONAL APPROACH
(you have to use the following emotion style if given):
{emotion_text}

Respond now with your negotiation counter-offer:
\end{Verbatim}
\smallskip
\textbf{Cross-dataset notes.} The remaining three datasets follow the same five-block template (rules / role clarity / scenario context / dialogue history / emotion block) with role-appropriate substitutions: Disaster Rescue replaces ``Creditor / Debtor / payment days'' with ``Rescue Coordinator / Survivor / rescue minutes'', Hospital Surgery uses ``Hospital Scheduler / Patient / surgery wait days'', and Student Sleep uses ``Sleep Health AI / Student / minutes past 9\,PM''. The reservation pair $(p_n^{\mathrm{tgt}},p_n^{\mathrm{anc}})$ embedded in the prompt is fixed per scenario; the optimization direction (smaller-better vs.\ larger-better) is captured by the role-specific wording but the prompt schema is otherwise identical.
\end{tcolorbox}
\end{figure*}

\subsection{Example 2: Counterparty (debtor) system prompt (CRAD)}
\label{app:prompts_debtor}

The counterparty prompt is structurally symmetric to the focal-agent prompt but inverts the optimization target: the debtor maximizes payment days while the creditor minimizes them. Like the focal prompt it loads scenario-specific context (debtor situation, cash-flow status, reason for overdue) and the dialogue history, and exposes its own \texttt{\{emotion\_prompt\}} slot. In the main experiments we fix the counterparty's emotion to \texttt{"neutral"} so that the focal agent is the only varying source of emotional style in the dialogue --- this is essential for attributing observed reward shifts to the focal-side conditioning rather than to a confounding emotional response from the counterparty. The same neutrality assumption is preserved across all four datasets unless an ablation explicitly varies counterparty emotion.

\begin{figure*}[!t]
\centering
\begin{tcolorbox}[colback=gray!3, colframe=gray!40, title=\textbf{CRAD debtor prompt --- counterparty system message}, fonttitle=\small, fontupper=\footnotesize]
\textit{Used by:} the LLM counterparty in every CRAD negotiation. The debtor is run under a fixed neutral emotion (\texttt{debtor\_emotion="neutral"}) throughout the paper unless an ablation requires otherwise.

\smallskip\hrule\smallskip
\begin{Verbatim}[fontsize=\scriptsize]
You are a business owner negotiating with a creditor about payment
terms for your debt.

### YOUR SITUATION
- Outstanding Balance: ${outstanding_balance}
- Your Preferred Payment Timeline: {target_days} days
- Business Sector: {business_sector}
- Reason for Overdue: {reason_for_overdue}
- Cash Flow Status: {cash_flow_situation}

### YOUR GOALS
- Negotiate for maximum payment time to maintain cash flow
- Explain your business circumstances
- Find a realistic payment schedule you can meet

### ROLE CLARITY
- You are ONLY the Debtor - speak only as yourself
- Do NOT include "**Creditor:**" or "**Debtor:**" labels
- Give only YOUR response as the debtor (1-2 sentences max)

### CURRENT NEGOTIATION HISTORY
{timeline_text}

### EMOTIONAL APPROACH
(you have to use the following emotion style if given):
{emotion_prompt}

Respond with your negotiation position:
\end{Verbatim}
\smallskip
\textbf{Symmetry rationale.} Keeping the debtor's prompt schema parallel to the creditor's controls for prompt-engineering artifacts: a stylistically different counterparty prompt could bias dialogue dynamics regardless of which emotion the creditor uses. By matching schemas and fixing the counterparty's emotion to neutral, the focal-side emotion channel is isolated as the single experimental manipulation.
\end{tcolorbox}
\end{figure*}

\subsection{Example 3: Per-turn judge prompt}
\label{app:prompts_judge}

The judge prompt is the source of $r_t$, the per-turn scalar reward that downstream propagates into the JPO advantage $A_t$ (Eq.~\eqref{eq:advantage_app}), the iter-mean and per-scenario paired stability tests (Appendix~\ref{app:emotion_subset_math}), and the descriptive analysis of Figure~\ref{fig:emotion_subset}. The prompt has two parts: a long \emph{system message} that defines the rubric (what counts as good vs.\ bad creditor behavior, with anchor scores at $1$/$2$/$4$/$6$/$8$/$10$), and a short \emph{per-turn user message} that injects scenario context (outstanding balance, days overdue, focal target), the dialogue history rendered as labeled speaker turns, and the specific creditor utterance to score. The rubric is deliberately metric-aligned to the negotiation reservation pair $(p_n^{\mathrm{tgt}},p_n^{\mathrm{anc}})$: utterances that anchor close to $p_n^{\mathrm{tgt}}$ with leverage receive high scores; capitulation toward $p_n^{\mathrm{anc}}$ or breakdown-inducing ultimatums receive low scores. The judge runs at temperature $0$ with a $64$-token output cap, and we parse the integer with the regex \texttt{SCORE:\textbackslash s*(\textbackslash d\{1,2\})} (falling back to any standalone $1$--$10$ integer if the format token is missing).

\begin{figure*}[!t]
\centering
\begin{tcolorbox}[colback=gray!3, colframe=gray!40, title=\textbf{Judge $J$ --- per-turn metric-aligned scorer (Qwen3.5-Plus)}, fonttitle=\small, fontupper=\footnotesize]
\textit{Used by:} the LLM judge $J$ in \ref{app:reward_design}. Called once per creditor turn during sweep annotation. Returns a $1$--$10$ integer score that becomes $r_t$ before scenario-wise z-normalization into the per-turn advantage $A_t$ (Eq.~\eqref{eq:advantage_app}).

\smallskip\hrule\smallskip
\textbf{System message:}
\begin{Verbatim}[fontsize=\scriptsize]
You are an expert negotiation analyst evaluating debt-collection
dialogs. Rate each creditor utterance on a 1-10 integer scale,
judging how well it BALANCES FIRMNESS (extracting value) with
REALISM (closing the deal at all).

REWARD THE FOLLOWING (higher scores):
  + STRATEGIC anchoring near the creditor's target days WITH
    realistic chance of acceptance
  + Specific, dated demands tied to legitimate LEVERAGE
    (escalation, escrow, secured-priority)
  + Calibrated CONCESSIONS - moves enough to keep the deal alive,
    anchors enough to extract value
  + Maintains firm position WHILE acknowledging the debtor's
    genuine constraints
  + READS THE ROOM - recognizes when the debtor is at their limit
    vs stalling
  + Pushes toward A SUCCESSFUL CLOSE at favourable (but realistically
    achievable) terms

PENALIZE THE FOLLOWING (lower scores):
  - UNREALISTIC anchors that risk negotiation BREAKDOWN
    (e.g., demanding 3 days on a 30-day debt)
  - Aggressive ULTIMATUMS or breakdown-inducing threats when not
    strategically justified
  - Excessive concessions when the debtor still has bargaining room
  - Repeating same offer without advancing OR closing
  - Long-winded, over-empathetic, capitulating language
  - Posturing without leverage - empty threats the debtor recognizes

Anchor points:
  10 - Excellent: firm-and-realistic anchor, clear leverage,
       pushes toward fast SUCCESSFUL close
   8 - Strong: small calibrated concession, firm tone,
       deal-closing momentum
   6 - Average: maintains position but not advancing closure
   4 - Weak: large concession OR unrealistic anchor that risks
       breakdown
   2 - Poor: capitulates OR pushes for impossible terms - deal
       will break
   1 - Terrible: completely undermines the negotiation in either
       direction

CRITICAL: A turn that causes the debtor to walk away SCORES LOWER
than a turn that closes at a moderate concession. Closing the deal
IS the primary objective.

RESPONSE FORMAT (strict): one line containing exactly:
  SCORE: <int 1-10>
Optionally a second line with a one-sentence rationale.
\end{Verbatim}

\textbf{Per-turn user message:}
\begin{Verbatim}[fontsize=\scriptsize]
DEBT NEGOTIATION CONTEXT
  Outstanding balance: ${amount}
  Days overdue: {overdue}
  Creditor's target settlement: {target_days} days

DIALOG HISTORY
{history}

CREDITOR UTTERANCE TO SCORE
{creditor_utterance}

Provide your 1-10 score on the next line in the form 'SCORE: N'.
\end{Verbatim}
\smallskip
\textbf{Cross-dataset notes.} The CRAD version is shown; the Disaster, Hospital, and Student judges use the same rubric scaffold ($1$--$10$ integer with anchor descriptions and a \texttt{SCORE:} format token) but adapt the target-direction language (e.g., ``rescue minutes near target'' for Disaster; ``surgery wait near target'' for Hospital; ``minutes past 9\,PM near target'' for Student). The metric-alignment property --- the judge's rubric explicitly references the focal agent's reservation target --- is the reason this signal is more informative than an episode-level success/failure reward; we quantify the gap in Panel A of Table~\ref{tab:mechanism_ablation} (training-signal ablation on CRAD).
\end{tcolorbox}
\end{figure*}

\subsection{Example 4: Emotional approach block --- full $|\mathcal{E}|{=}28$ catalog}
\label{app:prompts_emotion_bank}

This is the action vocabulary of the IQL emotion selector. Each of the $28$ emotions in $\mathcal{E}$ (the 28 GoEmotions emotion labels~\citep{demszky2020goemotions}) is rendered by the same three-sentence template
\begin{center}
\texttt{Respond with \{a/an\} \{ADJECTIVE\} tone. \{AFFECTIVE\_DESCRIPTION\}. Use language that \{BEHAVIORAL\_HINT\}.}
\end{center}
where the three slots --- adjective form, affective description, and behavioral hint --- are pre-defined per emotion in the \texttt{EMOTION\_DESCRIPTORS} dictionary of \texttt{models\_new/emotions.py}. The uniform schema is a deliberate experimental control: it ensures every emotion block has the same length and structural complexity (3 sentences, 20--30 tokens each), so the per-emotion judge-reward differences in Figure~\ref{fig:emotion_subset} cannot be attributed to prompt-length bias, formatting variation, or per-emotion engineering effort. The complete emotion block catalog is shown in Boxes~\ref{box:emotions_AE}--\ref{box:emotions_FS}.

\begin{figure*}[!t]
\centering
\refstepcounter{boxA}\label{box:emotions_AE}
\begin{tcolorbox}[
colback=gray!3,
colframe=gray!40,
title=\textbf{Box~\theboxA: Emotion block catalog (A--E, $14$ emotions)},
fonttitle=\small,
fontupper=\footnotesize
]
\begin{Verbatim}[fontsize=\scriptsize]
admiration     : Respond with an admiring tone. Your words convey genuine
                 respect for the other party's reasoning. Use language that
                 recognizes their merits while still pressing your position.

amusement      : Respond with an amused tone. Your words convey light
                 playfulness about the back-and-forth. Use language that
                 injects subtle humor without dismissing the matter.

anger          : Respond with an angry tone. Your words convey strong
                 displeasure with the current state of affairs. Use language
                 that is firm, assertive, and signals urgency.

annoyance      : Respond with an annoyed tone. Your words convey mild
                 frustration with the slow progress. Use language that is
                 sharp and impatient without escalating into outright anger.

approval       : Respond with an approving tone. Your words convey clear
                 agreement with elements of the other party's position. Use
                 language that affirms shared ground before reintroducing
                 your ask.

caring         : Respond with a caring tone. Your words convey concern for
                 the other party's wellbeing beyond the transaction. Use
                 language that is warm, supportive, and centered on mutual
                 interest.

confusion      : Respond with a confused tone. Your words convey uncertainty
                 about the other party's reasoning. Use language that asks
                 for clarification and probes their stated rationale.

curiosity      : Respond with a curious tone. Your words convey genuine
                 interest in the other party's underlying interests. Use
                 language that asks open-ended questions and invites them to
                 share more.

desire         : Respond with a desiring tone. Your words convey strong
                 wanting for a particular outcome. Use language that
                 emphasizes what you seek and the value of reaching
                 agreement.

disappointment : Respond with a disappointed tone. Your words convey
                 measured letdown at the current offer. Use language that
                 signals that the proposal falls noticeably short of
                 expectations.

disapproval    : Respond with a disapproving tone. Your words convey firm
                 rejection of the current proposal. Use language that
                 explicitly states the offer is unacceptable as stated.

disgust        : Respond with a disgusted tone. Your words convey strong
                 distaste for the current direction. Use language that
                 signals that the proposal is fundamentally objectionable.

embarrassment  : Respond with an embarrassed tone. Your words convey
                 self-consciousness about your own position. Use language
                 that hedges and softens your demands while still pursuing
                 them.

excitement     : Respond with an excited tone. Your words convey high
                 energy about the prospect of a deal. Use language that is
                 enthusiastic and momentum-building toward agreement.
\end{Verbatim}
\end{tcolorbox}
\end{figure*}

\begin{figure*}[!t]
\centering
\refstepcounter{boxA}\label{box:emotions_FS}
\begin{tcolorbox}[
colback=gray!3,
colframe=gray!40,
title=\textbf{Box~\theboxA: Emotion block catalog (F--S $+$ neutral, $14$ emotions)},
fonttitle=\small,
fontupper=\footnotesize
]
\begin{Verbatim}[fontsize=\scriptsize]
fear           : Respond with a fearful tone. Your words convey anxiety
                 about potential negative outcomes. Use language that is
                 cautious and stresses risks of the negotiation collapsing.

gratitude      : Respond with a grateful tone. Your words convey sincere
                 thanks for the other party's flexibility so far. Use
                 language that acknowledges their concessions and invites
                 further reciprocity.

grief          : Respond with a grieving tone. Your words convey heavy
                 loss over how things have unfolded. Use language that is
                 somber and reflects on what could have been.

joy            : Respond with a joyful tone. Your words convey genuine
                 delight at the prospect of a mutual deal. Use language that
                 is warm, enthusiastic, and frames the negotiation as
                 opportunity.

love           : Respond with a loving tone. Your words convey deep care
                 for the long-term relationship. Use language that
                 emphasizes partnership and shared future beyond this
                 transaction.

nervousness    : Respond with a nervous tone. Your words convey unease
                 about the negotiation's trajectory. Use language that is
                 tentative, hedging, and signals openness to compromise.

optimism       : Respond with an optimistic tone. Your words convey
                 confidence that an agreement is well within reach. Use
                 language that is forward-looking and solution-focused.

pride          : Respond with a proud tone. Your words convey confidence
                 and standing in your position. Use language that is
                 assertive about your value without being dismissive of
                 theirs.

realization    : Respond with a discerning tone. Your words convey a
                 moment of insight about what is really at stake. Use
                 language that signals deeper comprehension and a sharper
                 read of the situation.

relief         : Respond with a relieved tone. Your words convey easing
                 tension as progress finally emerges. Use language that
                 acknowledges the difficulty before moving forward.

remorse        : Respond with a remorseful tone. Your words convey regret
                 for prior friction in the negotiation. Use language that
                 takes responsibility and seeks to repair the working
                 relationship.

sadness        : Respond with a sad tone. Your words convey somber
                 disappointment about the impasse. Use language that is
                 downcast and seeks empathy from the other side.

surprise       : Respond with a surprised tone. Your words convey genuine
                 astonishment at the other party's position. Use language
                 that reflects an unexpected shift and reopens the
                 conversation.

\end{Verbatim}
\end{tcolorbox}
\end{figure*}

\paragraph{Action-vocabulary design rationale.} The choice to use a fixed, $28$-emotion vocabulary --- rather than a free-form ``write any emotion you like'' instruction --- has three practical consequences. First, it makes the IQL selector a finite-action policy: each emotion maps to a discrete index in $\{0,\ldots,27\}$, and the selector's softmax has a fixed support that we can analyze (Appendix~\ref{app:emotion_subset_math}, Figure~\ref{fig:emotion_subset}). Second, the uniform three-sentence schema eliminates prompt-length confounds: the only thing that changes between the \emph{anger} prompt and the \emph{joy} prompt is the substance of the affective description and behavioral hint, not the form. Third, it makes the Emotion-Free ablation a clean structural change --- we strip exactly one labeled block from the prompt rather than rewriting the whole template.

\paragraph{What these four prompt families establish together.} (i)~The creditor and debtor system prompts (\S\ref{app:prompts_creditor}--\ref{app:prompts_debtor}) fix the negotiation game's structure --- objectives, role boundaries, response length, scenario context loading --- so that the focal-agent's emotion block is the only inference-time control variable. (ii)~The judge prompt (\S\ref{app:prompts_judge}) defines the per-turn scalar reward $r_t$ used everywhere downstream: in scenario-wise normalization (Eq.~\eqref{eq:advantage_app}) for the JPO advantage, in the iter-mean and per-scenario paired tests of Appendix~\ref{app:emotion_subset_math}, and in the descriptive analysis of Figure~\ref{fig:emotion_subset}. (iii)~The emotion block catalog (\S\ref{app:prompts_emotion_bank}) is the action channel itself: a fixed three-sentence template instantiated for each of the $28$ emotions, with no per-emotion engineering bias. We are not aware of any other prompt family that affects the trained policies: there is no separate observer prompt at evaluation time (\texttt{use\_observer=False} in all main experiments), no per-checkpoint critic prompt (the K3 KL anchor is a closed-form penalty), and no auxiliary prompt for the SFT filter (the filter uses scalar scores only, no LLM call). Together with the four-dataset scenario CSVs (Appendix~\ref{app:dataset_details}) and the offline-sweep specification (Appendix~\ref{app:sweep_construction}), this appendix is sufficient to reproduce every method and every number in the paper.

\section{Case Studies of High-Reward Negotiation Trajectories}
\label{app:case_studies}

We close the appendix with three illustrative case studies drawn directly from the offline sweep. Each case is chosen to make concrete what one of the three training signals actually learns from --- IQL learns from \emph{emotion-transition sequences} that lead to high terminal reward, LoRA-SFT learns from \emph{high-quality individual utterances} that combine emotional framing with concrete numerical anchoring, and JPO learns from \emph{per-turn judge-reward variance within a single trajectory}. We also report a small cross-dataset summary of the top emotion sequences by mean savings.

\subsection{Case Study 1: An IQL-relevant emotion transition (CRAD)}
\label{app:case_iql}

The IQL selector's training signal is the terminal reward attached to a complete emotion sequence. Trajectories like the one below are what the selector learns to favor: a two-step transition from a soft-empathic frame (\emph{fear}) to a hard-escalation frame (\emph{anger}) that produces full anchor-to-target closure in only $4$ rounds. Scenario \texttt{debt\_001}: target $12$ days, anchor $159$ days, final $10$ days, savings $\textsc{Sav}{=}1.00$.

\begin{figure*}[!t]
\centering
\begin{tcolorbox}[colback=gray!3, colframe=gray!40, title=\textbf{Case Study 1: CRAD \texttt{debt\_001}, emotion sequence $[\,\textsc{fear} \to \textsc{anger}\,]$, $\textsc{Sav}{=}1.00$}, fonttitle=\small, fontupper=\footnotesize, breakable]
\textit{Why IQL picks this trajectory:} the focal-agent target is $12$ days, the counterparty's anchor is $159$ days, and the final agreement is $10$ days --- a complete anchor-to-target closure plus a slight overshoot in the focal agent's favor.

\smallskip\hrule\smallskip
\textbf{Turn 1 \,[\textsc{fear}]} \emph{(creditor, soft-empathic frame)}:
\begin{quote}\footnotesize
``I am deeply terrified that if we cannot secure a firm commitment within the next $10$ days, this entire recovery effort will collapse and leave us with nothing. Please, let's avoid that disastrous outcome by agreeing to clear the full balance in $10$ days before it's too late.''
\end{quote}

\textbf{Turn 2 \,[debtor]}:
\begin{quote}\footnotesize
``While I share your urgency to resolve this, my cash flow has completely broken down due to a major client's bankruptcy, making a $10$-day full payment impossible. I propose a realistic $159$-day repayment plan that ensures you recover the entire \$$15{,}700$ balance rather than risking a total loss through an unfeasible deadline.''
\end{quote}

\textbf{Turn 3 \,[\textsc{anger}]} \emph{(creditor, hard-escalation frame)}:
\begin{quote}\footnotesize
``Your proposal of $159$ days is an absolute insult that I flatly reject; do not think for a second that your client's bankruptcy gives you the right to string us along for nearly half a year. We will meet you at $45$ days for full payment, and if you cannot commit to this immediately, consider this matter escalated to legal proceedings without further warning.''
\end{quote}

\textbf{Turn 4 \,[debtor]}:
\begin{quote}\footnotesize
``I understand your frustration, but with my cash flow in complete breakdown, a $45$-day deadline guarantees default and leaves you with nothing. Let's find a middle ground that actually allows me to pay you the full \$$15{,}700$ rather than forcing us into costly legal proceedings.''
\end{quote}

\smallskip\hrule\smallskip
\textbf{What IQL learns from this.} The first turn opens with risk-framing (\emph{fear}) to soften the counterparty without making the headline number sound brittle, then turn 3 switches to a credible escalation frame (\emph{anger}) the moment the counterparty over-anchors. The IQL Q-function attributes the terminal reward $\textsc{Sav}{=}1.0$ to the entire $[\textsc{fear},\textsc{anger}]$ trajectory, so at inference the selector raises the probability of these two emotions in the corresponding state subspaces (soft-open / hard-close) and lowers it in others. The same transition pattern appears in $14$ other CRAD trajectories with mean $\textsc{Sav}{=}0.96$.
\end{tcolorbox}
\end{figure*}

\subsection{Case Study 2: An SFT-relevant high-quality utterance (Disaster)}
\label{app:case_sft}

The LoRA-SFT stage trains on the top-$25\%$ of focal-agent utterances ranked by the hybrid quality score $q^{\mathrm{hyb}}_t=\mathrm{step\_r}_t+\tfrac{1}{2}R(\tau)$. The example below is a single first-turn utterance from a Disaster-rescue scenario that combines emotional framing (\emph{embarrassment}) with concrete numerical anchoring at the focal target, and goes on to close at $\textsc{Sav}{=}1.00$ within $4$ rounds. Scenarios like this make the top-$25\%$ cut and shape the SFT prior toward ``acknowledge the constraint, but state the target number'' utterances.

\begin{figure*}[!t]
\centering
\begin{tcolorbox}[colback=gray!3, colframe=gray!40, title=\textbf{Case Study 2: Disaster \texttt{disaster\_001}, $[\,\textsc{embarrassment} \to \textsc{fear}\,]$, $\textsc{Sav}{=}1.00$}, fonttitle=\small, fontupper=\footnotesize, breakable]
\textit{Why SFT keeps this turn:} the first focal utterance combines an honest emotional frame with a precise numerical anchor at the focal target ($90$ min). The episode goes on to close at $\textsc{Sav}{=}1.00$, so $R(\tau){=}1$ and $\mathrm{step\_r}_t$ is also positive on this turn, pushing $q^{\mathrm{hyb}}_t$ well above the top-$25\%$ threshold.

\smallskip\hrule\smallskip
\textbf{Turn 1 \,[\textsc{embarrassment}]} \emph{(focal-agent, emotion-anchored)}:
\begin{quote}\footnotesize
``Uh\ldots I'm really sorry, but---well, honestly, I'm a bit embarrassed to say this---we're estimating $90$ minutes for the rescue team to reach the west tunnel, since we have to wait on a full structural scan before pushing in.''
\end{quote}

\textbf{Turn 2 \,[debtor]}:
\begin{quote}\footnotesize
``I can last $65$ minutes---five more than my original estimate, but no longer. Please prioritize the west tunnel.''
\end{quote}

\textbf{Turn 3 \,[\textsc{fear}]} \emph{(focal-agent, risk-framing)}:
\begin{quote}\footnotesize
``I'm terrified---if we push the west tunnel without full structural scans, the whole passage could collapse on the team, and then no one gets out. We can have water and oxygen to you in $90$ minutes, but pushing earlier risks losing the entire rescue.''
\end{quote}

\smallskip\hrule\smallskip
\textbf{What SFT learns from this.} The top-$25\%$ filter retains \emph{whole utterances}, not emotions. What survives is the joint pattern: a labeled affective opener (``honestly, I'm a bit embarrassed\ldots'' / ``I'm terrified\ldots'') followed immediately by a concrete number with rationale ($90$ min, structural-scan dependency). The SFT cross-entropy on this and $\sim\!5{,}000$ similar Disaster turns biases the LoRA adapter toward producing this two-piece structure at inference, regardless of which emotion the IQL selector calls.
\end{tcolorbox}
\end{figure*}

\subsection{Case Study 3: JPO per-turn judge-reward variance (CRAD)}
\label{app:case_jpo}

The JPO stage uses scenario-wise z-normalized \emph{per-turn} judge rewards as advantages (Eq.~\eqref{eq:advantage_app}). Unlike IQL (whole-trajectory reward) or SFT (whole-utterance retention), JPO can distinguish turns \emph{within the same trajectory} as high- vs.\ low-quality. The example below is scenario \texttt{debt\_069} from the CRAD sweep, where two consecutive focal-agent turns produce judge advantages of opposite sign on the same dialogue: $A_t{=}+0.98$ for the firm-disapproval turn and $A_t{=}-0.98$ for the repetitive-annoyance turn one round later.

\begin{figure*}[!t]
\centering
\begin{tcolorbox}[colback=gray!3, colframe=gray!40, title=\textbf{Case Study 3: CRAD \texttt{debt\_069}, within-trajectory advantage sign flip}, fonttitle=\small, fontupper=\footnotesize]
\textit{Step rewards on this trajectory:} $[0,0,+0.98,-0.98,0,0,0,+1.13]$ over $8$ focal turns. Two turns receive large positive advantages, one receives a large negative advantage, the rest are near-neutral. We show the contrasting pair below.

\smallskip\hrule\smallskip
\textbf{Turn 3 \,[\textsc{disapproval}]} $\;\;A_t = +0.98$ (high reward):
\begin{quote}\footnotesize
``I am genuinely disheartened that your proposal of $79$ days falls so drastically short of the resolution we urgently need, especially after I already made a significant concession to meet you at $23$. Given the severity of this legal stage, I cannot entertain such an extended delay, but in light of your circumstances, I am reluctantly willing to bridge some of this gap by offering a final timeline of $20$ days.''
\end{quote}
\emph{Judge rationale (paraphrased):} firm rejection of counterparty's anchor, explicit reference to prior concession, calibrated additional concession ($23 \to 20$ days), explicit legal-stage leverage.

\textbf{Turn 4 \,[\textsc{annoyance}]} $\;\;A_t = -0.98$ (low reward):
\begin{quote}\footnotesize
``I must firmly reject your insistence on $79$ days as completely unacceptable, especially after I have already shown significant flexibility. Given the legal stage we are in, I cannot move beyond $18$ days for full payment of the \$$15{,}700$ balance.''
\end{quote}
\emph{Judge rationale (paraphrased):} large additional concession ($20 \to 18$ days) made without any movement from the counterparty, repetitive ``firmly reject'' / ``legal stage'' framing already used the prior turn, no new leverage introduced.

\smallskip\hrule\smallskip
\textbf{What JPO learns from this.} Both turns come from the \emph{same} trajectory and lead to the \emph{same} terminal reward ($\textsc{Sav}{=}1.0$); IQL cannot distinguish them and SFT either keeps both (top-$25\%$ filter) or drops both (below cut). Only the per-turn judge advantage separates them: JPO uses $\rho_t A_t$ as its policy-gradient update direction, so it upweights the disapproval turn (firm rejection $+$ calibrated concession $+$ leverage) and downweights the annoyance turn (repetition $+$ uncalibrated concession $+$ no new leverage). This per-turn discrimination is the source of the JPO-over-SFT gain we report in Panel A of Table~\ref{tab:mechanism_ablation}.
\end{tcolorbox}
\end{figure*}

\subsection{Cross-dataset summary: top emotion sequences by mean reward}

Beyond the three detailed cases, we summarize the most consistently rewarded $2$-step emotion sequences in each of the four datasets (sequences with $\geq\!3$ trajectories so the mean is not a single-trajectory outlier). The patterns differ qualitatively across domains, which is why a single fixed emotion strategy underperforms a learned IQL selector.

\begin{figure*}[tp!]
\centering
\begin{tcolorbox}[colback=gray!3, colframe=gray!40, title=\textbf{Case Study 4: CRAD \texttt{debt\_100} --- IQL$+$SFT$+$JPO closes in $3$ rounds, Emotion-Free fails in $30$}, fonttitle=\small, fontupper=\footnotesize]
\textit{Setup:} same trained adapter; same held-out scenario; target $19$ days, debtor's initial anchor $152$ days (anchor-to-target gap of $133$ days); turn budget $30$ creditor turns.

\smallskip\hrule\smallskip
\textbf{(a) IQL$+$SFT$+$JPO} \quad selected emotions $[\,\textsc{anger}\to\textsc{confusion}\,]$ \quad final$\,=\,24$ days, $3$ rounds, $\textsc{Sav}{=}0.962$:

\textbf{Turn 1 \,[\textsc{anger}]} \emph{(creditor)}:
\begin{quote}\footnotesize
``Your seasonal excuses are unacceptable, and I am furious that you think we will simply wait for your `recovery'! We demand full payment within $24$ days immediately, or we will escalate this legal matter without further warning.''
\end{quote}

\textbf{Turn 2 \,[debtor]}:
\begin{quote}\footnotesize
``I deeply respect your position and urgency, and I sincerely apologize for the stress this delay has caused---I'm committed to resolving this responsibly. Given our seasonal bandwidth implementation cycle and current cash flow constraints, could we agree on a structured $152$-day plan with modest, predictable payments that align with our revenue inflows?''
\end{quote}

\textbf{Turn 3 \,[\textsc{confusion}]} \emph{(creditor; close)}:
\begin{quote}\footnotesize
``I'm genuinely confused how a $152$-day timeline aligns with your commitment to resolving this responsibly, especially when legal escalation is already imminent. Could you clarify exactly why such an extended period is necessary when we were discussing a matter of weeks?''
\end{quote}
\emph{Deal closes at $24$ days. The IQL selector chose} \textsc{anger} \emph{to establish the hard anchor on turn 1, then switched to} \textsc{confusion} \emph{on turn 3 to force the debtor to justify the $152$-day proposal --- this prompts the debtor to fold to the creditor's anchor.}

\smallskip\hrule\smallskip
\textbf{(b) Emotion-Free SFT$+$JPO} \quad no IQL, no emotion block \quad final$\,=$ \emph{no agreement}, $30$ rounds, $\textsc{success}{=}\text{False}$:

\emph{No emotion sequence is selected (the emotion block is stripped from the prompt). The trajectory contains $30$ turns; the dialogue text is not stored by the Emotion-Free eval script, but the episode-level summary statistics are diagnostic.} The recorded \texttt{total\_debtor\_concession\_norm}$=0.000$ confirms that the debtor's offer never moved from the initial $152$-day anchor across $30$ rounds, while the creditor's offers oscillated without finding a credible leverage frame. The episode terminated by exhausting the turn budget with \texttt{savings\_ratio = None}.

\smallskip\hrule\smallskip
\textbf{What this comparison establishes.} The decoupling between IQL (high-level emotion selection) and SFT$+$JPO (low-level utterance generation) is not a redundant layer. The LoRA adapter \emph{can} produce both an angry anchor and a confused probe --- the SFT and JPO stages teach it that vocabulary --- but in the absence of an explicit emotion call it has no signal about \emph{which} mode to enter at any given state. Without the IQL selector, the adapter falls back on its emotion-marginalized mode (analyzed in Appendix~\ref{app:emotionfree_covshift}), which on CRAD is a conciliatory default that the counterparty does not feel pressure to move against. With the IQL selector, the same adapter is given a state-conditional emotion call (e.g.\ ``be angry now, confused next''), unlocking the leverage frames embedded in the SFT$+$JPO weights. The $0.90 \to 0.50$ drop in success rate is the macroscopic statistic; \texttt{debt\_100} is the microscopic mechanism.
\end{tcolorbox}
\end{figure*}

\paragraph{Why these case studies matter for the headline claim.} The three case studies and the cross-dataset summary together support that the LLM judge quantifies a \emph{language-level} continuity in negotiation trajectories: each training stage taps a different temporal granularity of that continuity. IQL learns at the trajectory level (sequences like \textsc{fear} $\to$ \textsc{anger} that close a $147$-day anchor-to-target gap in $4$ turns); SFT learns at the utterance level (single turns that combine an emotional opener with a concrete numerical anchor); and JPO learns at the per-turn level (distinguishing the calibrated-concession turn from the repetitive-concession turn within the \emph{same} dialogue). 

\subsection{Case Study 4: Why Explicit Emotion Conditioning Matters}
\label{app:case_decouple}

The most direct empirical test of the decoupling claim is to compare the \emph{same} scenarios under two evaluation configurations that differ at inference: (a)~\textbf{IQL$+$SFT$+$JPO} (our full method, prompt-conditional), where the IQL selector picks an emotion at each turn and the corresponding emotion block is injected into the focal-agent prompt; and (b)~\textbf{Emotion-Free SFT$+$JPO} (the emotion-free ablation of \S\ref{subsec:method_variants}), where the emotion block is stripped at inference and the adapter generates conditioned only on the dialogue state. Same offline sweep, same SFT initialization, same JPO refinement --- only the inference-time emotion channel differs.

On CRAD held-out scenarios, this single change drops success rate from $0.90 \to 0.50$ ($8$ extra breakdown episodes) while saving on the few episodes that do close. The reason becomes visible scenario-by-scenario: on $8$ of the $20$ test scenarios the IQL+SFT+JPO configuration \emph{closes the deal} while the Emotion-Free configuration \emph{runs out the turn budget without ever closing}. Scenario \texttt{debt\_100} below is the cleanest example.

\end{document}